\pdfoutput=1
\documentclass[preprint]{elsarticle}

\usepackage[utf8]{inputenc}
\usepackage{moreverb,url}
\usepackage[,colorlinks,bookmarksopen,bookmarksnumbered,citecolor=red,urlcolor=red]{hyperref}
\usepackage{subcaption}
\usepackage{mathtools}
\usepackage{tikz}
\usepackage{xcolor}
\usetikzlibrary{positioning, calc, matrix, backgrounds}
\usepackage{amsfonts}
\usepackage{booktabs}
\usepackage{rotating}
\usepackage{placeins}

\newcommand{\secondreview}[1]{#1}

\newcommand{\added}[1]{#1}

\newcommand{\removed}[1]{}
\newcommand{\todo}[1]{}

\renewcommand{\vec}[1]{\boldsymbol{#1}}

\definecolor{blue}{HTML} {1EB1ED}
\definecolor{red}{HTML} {FC0D1B}
\definecolor{green}{HTML} {1AAF54}
\definecolor{yellow}{HTML} {FDBF2D}
\definecolor{purple}{HTML} {6F359E}
\definecolor{lightgreen}{HTML} {94CE58}

\newcommand{\tvspace}[1]{\renewcommand{\arraystretch}{#1}}
\newcommand{\thspace}[1]{\renewcommand{\tabcolsep}{#1}}

\begin{document}

\title{Learning a bidirectional mapping between human whole-body motion and natural language using deep recurrent neural networks}

\author[]{Matthias Plappert\corref{cor}}
\ead{matthias.plappert@partner.kit.edu}
\author[]{Christian Mandery}
\ead{mandery@kit.edu}
\author[]{Tamim Asfour}
\ead{asfour@kit.edu}

\cortext[cor]{Corresponding author}
\address{High Performance Humanoid Technologies (H\textsuperscript{2}T),
Karlsruhe Institute of Technology (KIT),
Adenauerring 2 (Building 50.20),
76131 Karlsruhe,
Germany
}





\begin{abstract}
Linking human whole-body motion and natural language is of great interest for the generation of semantic representations of observed human behaviors as well as for the generation of robot behaviors based on natural language input.
While there has been a large body of research in this area, most approaches that exist today require a symbolic representation of motions (e.g. in the form of motion primitives), which have to be defined a-priori or require complex segmentation algorithms.
In contrast, recent advances in the field of neural networks and especially deep learning have demonstrated that sub-symbolic representations that can be learned end-to-end usually outperform more traditional approaches, for applications such as machine translation.
In this paper we propose a generative model that learns a bidirectional mapping between human whole-body motion and natural language using \emph{deep recurrent neural networks} (RNNs) and \emph{sequence-to-sequence learning}.
Our approach does not require any segmentation or manual feature engineering and learns a distributed representation, which is shared for all motions and descriptions.
We evaluate our approach on 2\,846~human whole-body motions and 6\,187~natural language descriptions thereof from the \emph{KIT Motion-Language Dataset}.
Our results clearly demonstrate the effectiveness of the proposed model:
We show that our model generates a wide variety of realistic motions only from descriptions thereof in form of a single sentence.
Conversely, our model is also capable of generating correct and detailed  natural language descriptions from human motions.
\end{abstract}


\maketitle

\section{Introduction}
\label{sec:intro}
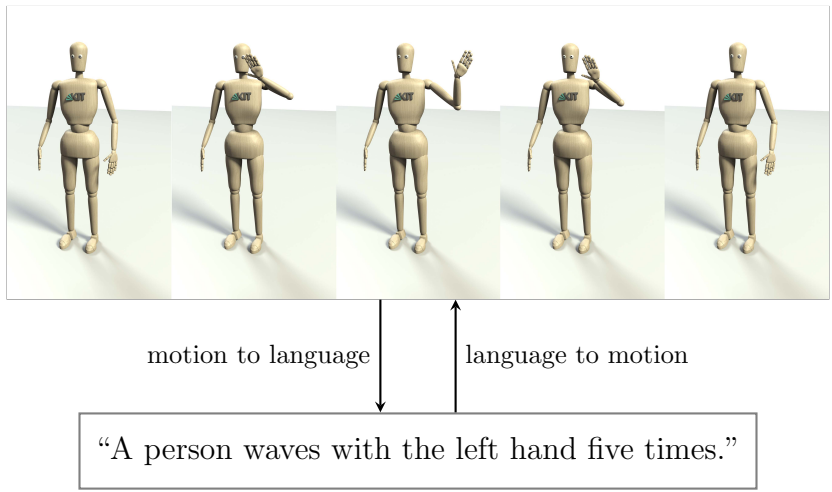
\begin{figure}[h]
    \centering
    \begin{tikzpicture}[>=stealth, thick]
        \tikzstyle{box} = [rectangle,draw=gray]

        \node[box, inner sep=0pt] (motion) {\includegraphics[width=0.9\textwidth]{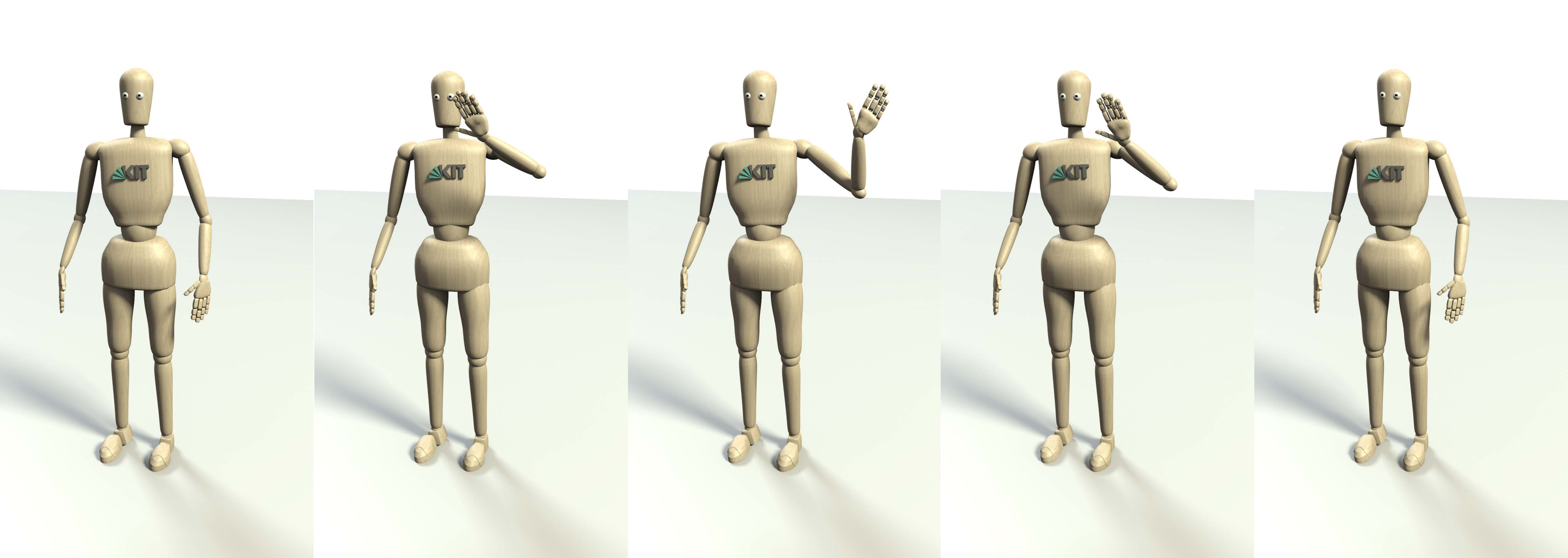}};
        \node[box,below=1.5cm of motion, minimum height=1cm, minimum width=9cm] (language) {\large``A person waves with the left hand five times.''};
        \draw[->, thick] ([xshift=-0.5cm]motion.south) -- ([xshift=-0.5cm]language.north)node[midway,left] {motion to language};
        \draw[<-, thick] ([xshift=0.5cm]motion.south) -- ([xshift=0.5cm]language.north)node[midway,right] {language to motion};
    \end{tikzpicture}
    \caption{Illustration of the desired bidirectional mapping between human whole-body motion (top) and natural language (bottom).}
    \label{fig:intro}
\end{figure}

An intriguing way to instruct a robot is to first demonstrate the task at hand.
In such a setup, a human teacher performs the necessary steps while the robot observes the human's motion.
This way of robot programming is commonly referred to as \emph{programming by demonstration}~\citep{kuniyoshi1994, dillmann2000learning, DBLP:reference/robo/BillardCDS08} and has been extensively studied.
However, observing only the motion of a human teacher is often not sufficient as the demonstrator will often include additional or corrective instructions to the student using natural language.
In other words, the teacher-student interaction is inherently multi-modal.

Natural language presents itself as an intuitive way of communicating with the robot since it can be used to describe even rather complex motions and their parameterizations.
For example, the description \emph{``A person waves with the left hand five times.''} encodes the motion (\emph{waving}), the body part that should perform it (\emph{left hand}) and the number of repetitions (\emph{five times}).
Enabling a robot to combine such rich descriptions in natural language with human whole-body motion therefore facilitate a much richer human-robot communication.

In recent years, deep learning~\citep{lecun2015deep, Goodfellow-et-al-2016-Book} has proven to be very successful in computer vision~\citep{DBLP:conf/nips/KrizhevskySH12, DBLP:journals/corr/HeZRS15}, natural language processing~\citep{seq2seq1, DBLP:conf/icml/GlorotBB11, DBLP:journals/corr/WuSCLNMKCGMKSJL16} and speech recognition~\citep{DBLP:conf/icassp/GravesMH13}.
More recently, researchers have also reported promising results when applying deep learning techniques to problems in robotics~\citep{DBLP:journals/corr/LevinePKQ16, DBLP:journals/corr/GuHLL16, DBLP:journals/corr/LevineFDA15}.

In this paper, we use deep learning techniques to link human whole-body motion and natural language. 
More specifically, we make use of \emph{sequence-to-sequence learning}~\citep{seq2seq1} to learn a bidirectional mapping between human whole-body motion and natural language.
Human whole-body motion is represented in joint space under the \emph{Mater Motor Map}~(MMM)~\citep{DBLP:conf/humanoids/TerlemezUMDVA14} framework and descriptions thereof are in the form of complete English sentences.
\autoref{fig:intro} illustrates the desired mapping.

On one hand, this mapping allows us to generate rich descriptions of observed human motion, which can, for example, be used in a motion database.
On the other hand, our model is capable of generating a wide range of different motions only from a description thereof in natural language.
Even more so, the proposed system is capable of successfully synthesizing certain variations of motion, e.g. waving with the left or the right hand as well as walking quickly or slowly simply by specifying this parametrization in the natural language description.

The remainder of this paper is organized as follows.
In \autoref{sec:related} we review work that is related to our approach of combining human motion and natural language.
\autoref{sec:representation} describes in detail how we represent both modalities, human motion and natural language, for use in the proposed bidirectional mapping.
The model that is used to learn this mapping is presented in \autoref{sec:model}.
In \autoref{sec:evaluation} we show that the proposed approach is capable of learning the desired bidirectional mapping.
We also analyze the model and its learned representations in depth.
Finally, \autoref{sec:conclusion} summaries and discusses our results and points out promising areas for future work.

\section{Related work}
\label{sec:related}
Different models to encode human motion have been proposed in the literature.
\cite{DBLP:conf/icra/TakanoYSYN06}, \cite{DBLP:journals/ijrr/KulicTN08} and \cite{DBLP:conf/humanoids/HerzogUK08} use \emph{hidden Markov models}~(HMMs) to learn from observation.
Their model can also be used to generate motion sequences by sampling from it.
\cite{DBLP:conf/nips/TaylorHR06} and \cite{DBLP:conf/icml/TaylorH09} propose \emph{conditional restricted Boltzmann machines} (CRBMs) to learn from and then generate human whole-body motion.
\cite{DBLP:journals/tsmc/CalinonGB07} use a \emph{Gaussian mixture model} (GMM) to encode motion data.
A different approach is proposed by \cite{Schaal2006}, who uses \emph{dynamic movement primitives} (DMPs) to model motion using a set differential equations.
More recently, \cite{DBLP:journals/corr/FragkiadakiLM15} and \cite{DBLP:journals/corr/JainZSS15} have used \emph{recurrent neural networks} (RNNs) to learn to generate human motion from observation.
\cite{DBLP:conf/nips/MordatchLAPT15} have combined \emph{trajectory optimization} and \emph{deep neural networks} to generate complex and goal-directed movement for a diverse set of characters.\\
\cite{DBLP:journals/corr/ButepageBKK17} use deep feed-forward neural networks and an encoder-decoder architecture to learn a lower-dimensional latent representation that can be used for classification and motion generation.

While many models have been proposed to encode human motion, less research has been conducted on the question how to combine human motion and natural language.
\cite{DBLP:conf/iros/TakanoKN07} and \cite{DBLP:journals/ras/TakanoN15} describe a system that allows to learn a mapping between human motion and word labels.
The authors segment human motion and encode the resulting motion primitives into hidden Markov models, which then form the \emph{motion symbol space} or \emph{proto symbol space}.
Similarly, the \emph{word space} is constructed from the associated word labels.
Finally, the authors describe a projection between the motion symbol space and word space, which allows to obtain a sequence of motion symbols from a sequence of words and vice versa.

\cite{DBLP:conf/humanoids/TakanoN08, DBLP:conf/icra/TakanoN09, DBLP:conf/icra/TakanoN12, DBLP:journals/ijrr/TakanoN15} learn a bidirectional mapping between human motion and natural language in the form of a complete sentence.
The authors propose two components that, when combined, realize the desired mapping.
In the \emph{motion language model}, motion primitives, which are encoded into HMMs, are probabilistically related to words using latent variables.
These latent variables represent non-observable properties like the semantics.
The conditional probabilities that govern the association between motion and language are obtained by using the EM algorithm.
The second part, the \emph{natural language model}, captures the syntactical structure of natural language.
Different approaches to realize this model have been described by the authors, e.g. HMMs~\citep{DBLP:conf/humanoids/TakanoN08, DBLP:conf/icra/TakanoN09} or bigram models~\citep{DBLP:conf/icra/TakanoN12, DBLP:journals/ijrr/TakanoN15}.
Both models are finally combined to generate natural language descriptions of motion by first recognizing the motion, then obtaining an unordered set of likely words associated with that motion (using the motion language model) and finally finding a likely sequence of these words (using the natural language model).
This approach is commonly referred to as \emph{bag-of-words}.
Similarly, the most likely motion can be obtained from a description thereof in natural language by searching for the motion symbol with maximum likelihood given the word sequence.

\cite{DBLP:conf/ro-man/MedinaSLTH12} present an approach where motion symbols in the form of motion primitives are learned using \emph{parametric hidden Markov models} (PHMMs).
Adverbs (e.g. \emph{slowly}) are used to parametrize the PHMMs and the natural language is modeled in a similar way to aforementioned works using a bigram language model.
Although their approach can enable the generation of textual representations from motions, the presented evaluation only covers the other direction of motion generation from textual descriptions.
This evaluation considers a rather limited set of 7 different motion primitives and 13 words, and is based on a virtual scenario of a human-robot cooperation task with a 2 degrees of freedom haptic interface.
A different approach is described by \cite{DBLP:journals/adb/SugitaT05} and \cite{DBLP:conf/icra/OgataMTKO07, DBLP:conf/iros/OgataMTKO07}  where the authors use a \emph{recurrent neural network with parametric bias} (RNNPB) model to combine movement of simple robots (e.g. a robot platform or a robot arm) with simple commands in the form of words (e.g. \emph{push red}).
The authors demonstrate that their approach can generate the appropriate trajectories corresponding to a given command and vice versa.
Attempts have been made to gradually increase the complexity and supported variety of the commands~\citep{DBLP:conf/icann/ArieEJLST10, DBLP:conf/riiss/OgataO13}.

In recent years, there has been extensive work in the field of deep learning to combine natural language with other modalities like images~\citep{DBLP:conf/cvpr/KarpathyL15, DBLP:conf/cvpr/VinyalsTBE15} and videos~\citep{DBLP:conf/cvpr/DonahueHGRVDS15, DBLP:conf/naacl/VenugopalanXDRM15}. A recent survey on proposed solutions to the problem of visualizing natural language descriptions is given in \cite{natural_language_descr_survey}.

\section{Data representation}
\label{sec:representation}
\subsection{Human whole-body motion}
\label{sec:human-motion}
In this work, we only consider human whole-body motion that has been recorded with an optical marker-based motion capture system.
Briefly speaking, such a system observes a set of reflective markers placed on specific anatomical landmarks of the human subject using multiple cameras that are positioned around the subject.
The Cartesian coordinates of each marker can then be reconstructed using triangulation.
For an in-depth discussion of motion capture techniques, we refer the reader to~\cite{field2011mocap}.

While this approach allows for highly accurate motion acquisition, the resulting representation of motion in the form of marker trajectories has several drawbacks.
First, the positions of the markers depend on the reference coordinate system, which varies across recordings and thus would require some sort of normalization to obtain an invariant representation.
Second, different marker sets with a varying number of markers or different marker locations on the human body may be used.
Third, the data is high-dimensional since each marker position requires three dimensions and usually more than $50$ markers are used.

\begin{figure}[h]
    \centering
        \includegraphics[width=0.5\textwidth]{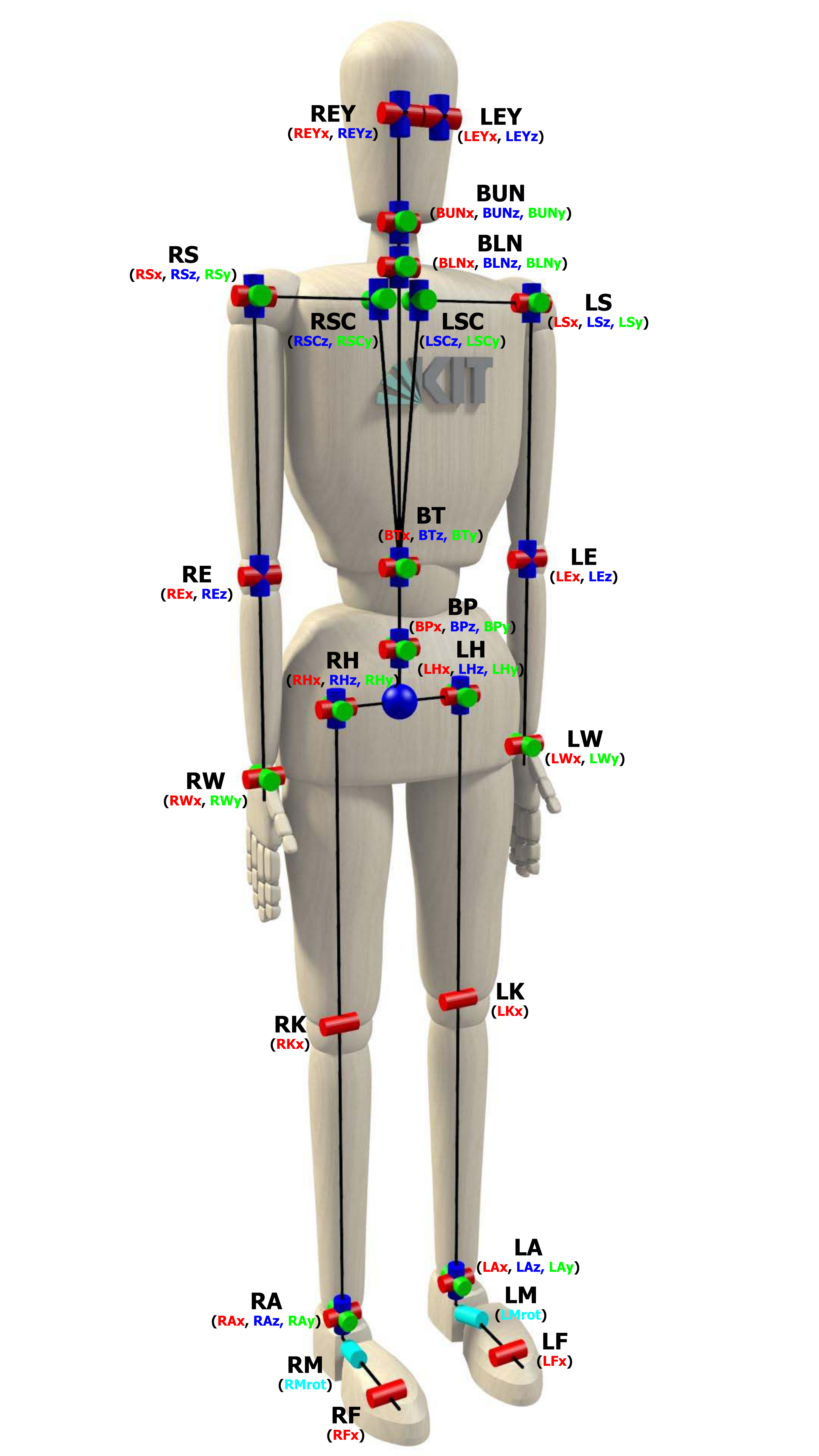}
    \caption{The kinematic model of the \emph{Master Motor Map} (MMM) framework \citep{DBLP:journals/trob/ManderyTDVA16}.}
    \label{fig:mmm}
\end{figure}

We therefore use the \emph{Master Motor Map}~(MMM) framework~\citep{DBLP:journals/trob/ManderyTDVA16, DBLP:conf/humanoids/TerlemezUMDVA14, DBLP:conf/icra/AzadAD07} to represent human whole-body motion in \emph{joint space}.
This is realized by the MMM reference model, which specifies the kinematics of the human body (see \autoref{fig:mmm}).
The conversion from Cartesian to joint space is then achieved by minimizing the squared distance between the physical markers on the human subject and the virtual markers on the reference model w.r.t. to the joint angles of the kinematic model.
The resulting joint representation is no longer dependent on a reference coordinate system, abstracts away the concrete marker set that was used during recording and has significantly lower dimensionality than the Cartesian representation.

Throughout this paper, we use $J=44$~joints of the MMM reference model to represent human motion, which are distributed over the torso, arms and legs of the model.
The remainder of the degrees of freedom that the model features (e.g. individual fingers and eyes) are not used since they are less important for this work and also hard to track.

We also introduce an additional binary feature which is enabled as long as the motion is ongoing.
This feature is necessary for two reasons:
First, we pad all sequences to have equal length due to implementation details and the binary flag indicates the active part of the motion.
Second and more importantly, the generative part of our proposed model will also predict the binary flag and thus can be used to indicate if the generation is still ongoing or if it has finished.

More formally, each motion $\vec{M}$ is thus represented as a sequence of length $N$
\begin{equation}
    \vec{M} = \left(\vec{m}^{(1)}, \vec{m}^{(2)}, \ldots, \vec{m}^{(t)}, \ldots, \vec{m}^{(N)}\right),
\end{equation}
where each timestep $\vec{m}^{(t)}$ is defined as
\begin{equation}
    \vec{m}^{(t)} \in \mathbb{R}^{J} \times \{0, 1\}.
\end{equation}
\added{For the padded part of the motion, we set the first $J$ elements to zero.}

Furthermore, we scale the individual joints to have zero mean and variance one.
We also down-sample each motion from $100$~Hz to $10$~Hz, which results in shorter sequences that are less resource-expensive during training and evaluation of our model.
However, we do not discard the remaining motion data since we split each original motion sequence into $10$~down-sampled sequences by applying a variable offset ${t_0 \in \{0, \ldots, 9\}}$.
We then treat these as additional data during training, effectively introducing noise into the training process.
Since we do this for all sequences, the original distribution of different motion types in the dataset is maintained.

\begin{figure}[h]
    \centering
    \includegraphics[width=0.8\textwidth]{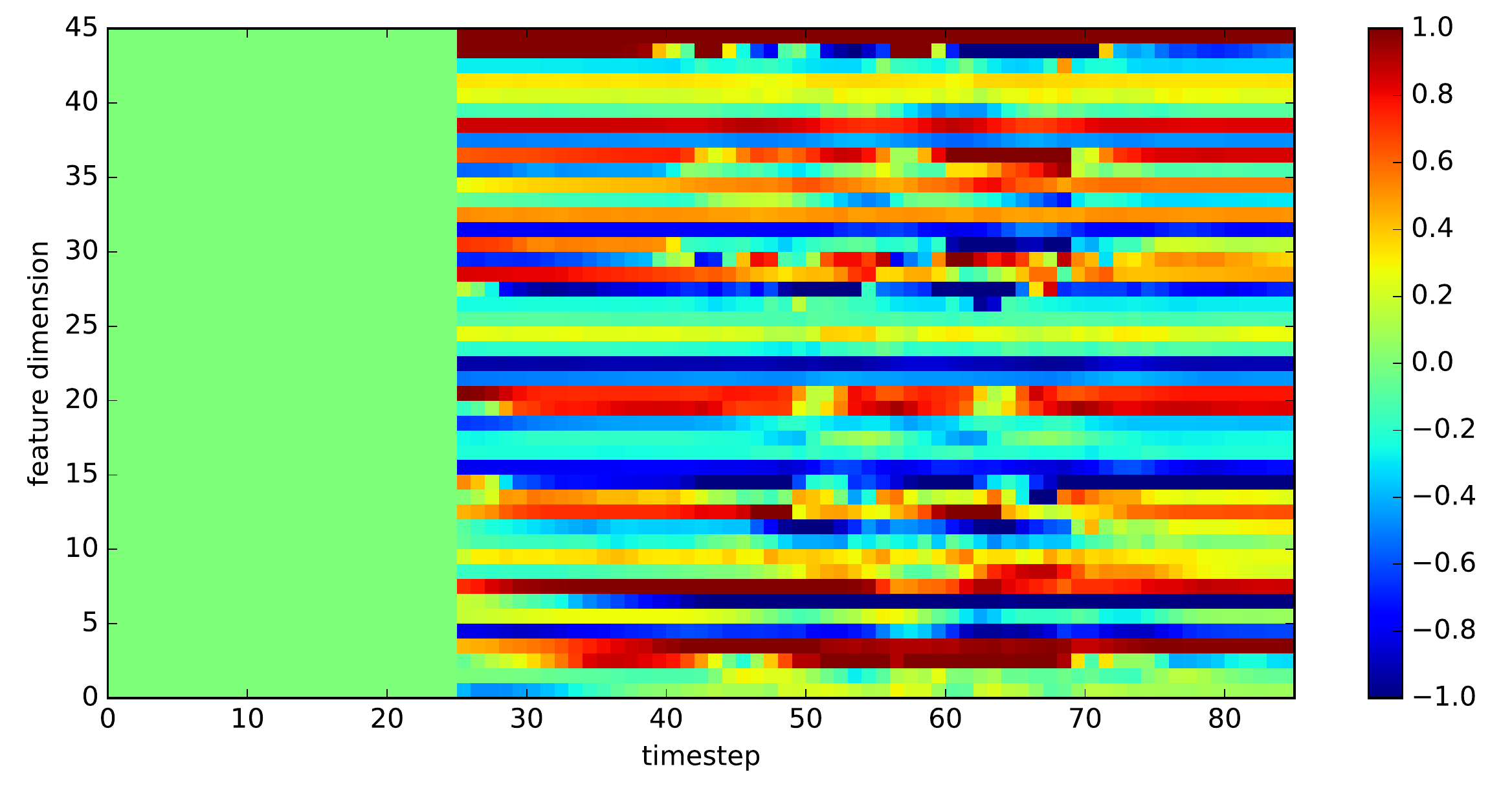}
    \caption{Representation of an exemplary human motion in which the subject walks forward. The first $J=44$ dimensions are the joint values of the MMM model and dimension $J+1=45$ is the binary active feature, plotted over time. The motion is padded, which explains the constant values for the first 25~timesteps. Notice that the motion is not active during this part, as indicated by the binary feature.}
    \label{fig:motion}
\end{figure}

\autoref{fig:motion} illustrates the described representation of a single human whole-body motion.

\subsection{Natural language descriptions}
We represent the natural language annotations on the word-level.
More concretely, we first normalize by transforming each sentence to lower case letters, remove all punctuation and apply minor spelling corrections for commonly misspelled words.
We also pad all sequences to have equal length by introducing a special \texttt{PAD} word and, similarly, use special \texttt{SOS} (start of sentence) and \texttt{EOS} (end of sentence) words to mark the start and end of a sentence, respectively.
Next, we tokenize each sentence into individual words and assign each word a unique integer index.

More formally, each sentence $\vec{w}$ is thus represented as a sequence of length $M$
\begin{equation}
    \vec{w} = \left(w^{(1)}, w^{(2)}, \ldots, w^{(t)}, \ldots, w^{(M)}\right),
\end{equation}
where each word is represented as an integer ${w^{(t)} \in \mathbb{N}}$.
Note that, typically, ${M \ll N}$ (recall that $N$ denotes the motion sequence length), i.e. the motion and language sequences are not required to have equal length.

In practice, we encode each word using one-hot encoding $\vec{\bar{w}}^{(t)} \in \{0,1\}^V$ over a vocabulary of size~$V$ instead of using the integer representation (since this would imply some order):
\begin{equation}
  \bar{w}^{(t)}_i = \left\{
  \begin{array}{@{}ll@{}}
    1, & \text{if}\ i=w^{(t)} \\
    0, & \text{otherwise}
  \end{array}\right.
\end{equation}
However, this representation has the problem that its dimensionality grows linearly with $V$ (the number of words in the vocabulary).
\emph{Word embeddings}~\citep{word_embeddings1} resolve this problem by projecting the one-hot encoded word to a continuous, but much lower-dimensional \emph{embedding space}.
Depending on the training procedure, these embeddings have also been shown to group semantically similar words closer together.

In this work, we learn this projection end-to-end when training the entire model.
However, in future work we might consider to use the weights of an embedding layer that was pre-trained on a large text corpus, e.g. using the open source \texttt{word2vec} implementation.\footnote{\url{https://code.google.com/archive/p/word2vec/}}

\section{Model}
\label{sec:model}
We describe the two directions of the mapping between human motion and natural language separately.
However, since both models share a common approach commonly referred to as \emph{sequence-to-sequence learning}~\citep{seq2seq1}, we explain this shared mode of operation before describing each model individually.

\begin{figure}[h]
    \centering
    \begin{subfigure}{0.49\textwidth}
        \resizebox{\textwidth}{!}{%
    \begin{tikzpicture}[node distance=1.5em and 1em, auto, minimum width=3.5em, minimum height=1.5em, >=stealth, thick]

        \tikzset{font=\footnotesize}
        \tikzstyle{input} = [rectangle]
        \tikzstyle{output} = [rectangle]
        \tikzstyle{encoder_rnn} = [text width=3em,align=center, rectangle, draw=yellow, fill=yellow!50, minimum height=3em]
        \tikzstyle{embedding} = [rectangle, draw=blue, fill=blue!50]
        \tikzstyle{decoder_rnn} = [rectangle, draw=green, fill=green!50, minimum height=3em]
        \tikzstyle{decoder_fc} = [rectangle, draw=lightgreen, fill=lightgreen!50]
        \tikzstyle{decoder} = [rectangle, draw=red, fill=red!50]
        \tikzstyle{placeholder} = [rectangle, minimum width=1.5em]
        \tikzstyle{label} = [rectangle, minimum width=0]
        \tikzstyle{context} = [circle, draw=purple, fill=purple!50, minimum width=0]
        \tikzstyle{embedding_arrow} = []

        %
        %

        \node[decoder] (decoder0) {Decoder};
        \node[decoder, right= of decoder0] (decoder1) {Decoder};
        \node[placeholder, right= of decoder1] (decoder2) {$\ldots$};

        \node[decoder_fc, below= of decoder0] (decoder_fc0) {FC};
        \node[decoder_fc, below= of decoder1] (decoder_fc1) {FC};
        \node[placeholder, below= of decoder2] (decoder_fc2) {$\ldots$};

        \node[decoder_rnn, below=1em of decoder_fc0] (decoder_rnn0) {RNNs};
        \node[decoder_rnn, below=1em of decoder_fc1] (decoder_rnn1) {RNNs};
        \node[placeholder, minimum height=3em, below=1em of decoder_fc2] (decoder_rnn2) {$\ldots$};

        \node[embedding, below= of decoder_rnn0] (decoder_embedding0) {Embed};
        \node[embedding, below= of decoder_rnn1] (decoder_embedding1) {Embed};
        \node[placeholder, below= of decoder_rnn2] (decoder_embedding2) {$\ldots$};

        \node[input, below= of decoder_embedding0] (sos) {\texttt{SOS}};

        \node[context, left=1em of decoder_rnn0] (context) {$\vec{c}$};

        \node[output, above= of decoder0] (w1) {$\hat{w}^{(1)}$};
        \node[output, above= of decoder1] (w2) {$\hat{w}^{(2)}$};
        \node[placeholder, above= of decoder2] (wt) {$\ldots$};

        \draw[->] (decoder0.north) -- (w1.south);
        \draw[->] (decoder1.north) -- (w2.south);
        \draw[->] (sos.north) -- (decoder_embedding0.south);

        \draw[->,draw=blue] (decoder_embedding0.north) -- (decoder_rnn0.south);
        \draw[->,draw=blue] (decoder_embedding1.north) -- (decoder_rnn1.south);
        
        \draw[->,draw=black] (decoder_rnn0.north) -- (decoder_fc0.south);
        \draw[->,draw=black] (decoder_rnn1.north) -- (decoder_fc1.south);

        \draw[->,draw=black] (decoder_fc0.north) -- (decoder0.south)node[label,midway,right] {$\vec{\hat{y}}^{(1)}$};
        \draw[->,draw=black] (decoder_fc1.north) -- (decoder1.south)node[label,midway,right] {$\vec{\hat{y}}^{(2)}$};

        \coordinate[below left=0.75em and 0 of decoder_rnn0] (anchor_c0);
        \draw[-, draw=purple] (context.south) |- (anchor_c0);
        \draw[->, draw=purple] (anchor_c0) -| ([xshift=-0.5em]decoder_rnn0.south);

        \coordinate[below left=0.75em and 0 of decoder_rnn1] (anchor_c1);
        \draw[-, draw=purple] (anchor_c0) |- (anchor_c1);
        \draw[->, draw=purple] (anchor_c1) -| ([xshift=-0.5em]decoder_rnn1.south);

        \coordinate[below left=0.75em and 0 of decoder_rnn2] (anchor_c2);
        \draw[->, draw=purple] (anchor_c1) |- (anchor_c2);

        \coordinate[above right=0.5em and 0.5em of decoder0] (anchor_w10) {};
        \coordinate[below right=0.5em and 0.5em of decoder_embedding0] (anchor_w11) {};
        \draw[-] (decoder0.north) |- (anchor_w10);
        \draw[-] (anchor_w10) -- (anchor_w11);
        \draw[->] (anchor_w11) -| (decoder_embedding1.south);

        \coordinate[above right=0.5em and 0.5em of decoder1] (anchor_w20) {};
        \coordinate[below right=0.5em and 0.5em of decoder_embedding1] (anchor_w21) {};
        \draw[-] (decoder1.north) |- (anchor_w20);
        \draw[-] (anchor_w20) -- (anchor_w21);
        \draw[->] (anchor_w21) -| (decoder_embedding2.south);

        \draw[->,draw=green] (decoder_rnn0.east) -- (decoder_rnn1.west);
        \draw[->,draw=green] (decoder_rnn1.east) -- (decoder_rnn2.west);

        %
        %

        \node[encoder_rnn, left=3.5em of decoder_rnn0] (encoder_rnnN) {BRNNs};
        \node[encoder_rnn, left= of encoder_rnnN] (encoder_rnnt) {BRNNs};
        \node[encoder_rnn, left= of encoder_rnnt] (encoder_rnn1) {BRNNs};

        \node[input, below=4.5em of encoder_rnnN] (mN) {$\vec{m}^{(N)}$};
        \node[input, below=4.5em of encoder_rnnt] (mt) {$\ldots$};
        \node[input, below=4.5em of encoder_rnn1] (m1) {$\vec{m}^{(1)}$};

        \draw[->,draw=yellow] ([yshift=+0.5em]encoder_rnn1.east) -- ([yshift=+0.5em]encoder_rnnt.west);
        \draw[->,draw=yellow] ([yshift=+0.5em]encoder_rnnt.east) -- ([yshift=+0.5em]encoder_rnnN.west);
        \draw[<-,draw=yellow] ([yshift=-0.5em]encoder_rnn1.east) -- ([yshift=-0.5em]encoder_rnnt.west);
        \draw[<-,draw=yellow] ([yshift=-0.5em]encoder_rnnt.east) -- ([yshift=-0.5em]encoder_rnnN.west);
        \draw[->,draw=yellow] ([yshift=+0.5em]encoder_rnnN.east) -| (context.west);
        
        \coordinate[above left=0.5em and 0.5em of encoder_rnn1] (anchor_context_backward);
        \draw[-,draw=yellow] ([yshift=-0.5em]encoder_rnn1.west) -| (anchor_context_backward);
        \draw[->,draw=yellow] (anchor_context_backward) -| (context.north);

        \draw[->,draw=black] (m1.north) -- (encoder_rnn1.south);
        \draw[->,draw=black] (mt.north) -- (encoder_rnnt.south);
        \draw[->,draw=black] (mN.north) -- (encoder_rnnN.south);
    \end{tikzpicture}
}
        \caption{Motion-to-language model.}
        \label{fig:m2l}
    \end{subfigure}
    \hfill
    \begin{subfigure}{0.49\textwidth}
        \resizebox{\textwidth}{!}{%
  \begin{tikzpicture}[node distance=1.5em and 1em, auto, minimum width=3.5em, minimum height=1.5em, >=stealth, thick]
      \tikzset{font=\footnotesize}
      \tikzstyle{input} = [rectangle]
      \tikzstyle{output} = [rectangle]
      \tikzstyle{encoder_rnn} = [text width=3em,align=center, rectangle, draw=yellow, fill=yellow!50, minimum height=3em]
      \tikzstyle{embedding} = [rectangle, draw=blue, fill=blue!50]
      \tikzstyle{decoder_rnn} = [rectangle, draw=green, fill=green!50, minimum height=3em]
      \tikzstyle{decoder_fc} = [rectangle, draw=lightgreen, fill=lightgreen!50]
      \tikzstyle{decoder} = [rectangle, draw=red, fill=red!50]
      \tikzstyle{placeholder} = [rectangle, minimum width=1.5em]
      \tikzstyle{label} = [rectangle, minimum width=0]
      \tikzstyle{context} = [circle, draw=purple, fill=purple!50, minimum width=0]
      \tikzstyle{embedding_arrow} = []

      %
      %

      \node[decoder] (decoder0) {Decoder};
      \node[decoder, right= of decoder0] (decoder1) {Decoder};
      \node[placeholder, right= of decoder1] (decoder2) {$\ldots$};

      \node[decoder_fc, below= of decoder0] (decoder_fc0) {FC};
      \node[decoder_fc, below= of decoder1] (decoder_fc1) {FC};
      \node[placeholder, below= of decoder2] (decoder_fc2) {$\ldots$};

      \node[decoder_rnn, below=1em of decoder_fc0] (decoder_rnn0) {RNNs};
      \node[decoder_rnn, below=1em of decoder_fc1] (decoder_rnn1) {RNNs};
      \node[placeholder, below=1em of decoder_fc2, minimum height=3em] (decoder_rnn2) {$\ldots$};

      \node[input, below=4.5em of decoder_rnn0] (sos) {$[1, \dots, 1]$};

      \node[context, left=1em of decoder_rnn0] (context) {$\vec{c}$};

      \node[output, above= of decoder0] (w1) {$\vec{\hat{m}}^{(1)}$};
      \node[output, above= of decoder1] (w2) {$\vec{\hat{m}}^{(2)}$};
      \node[placeholder, above= of decoder2] (wt) {$\ldots$};

      \draw[->] (decoder0.north) -- (w1.south);
      \draw[->] (decoder1.north) -- (w2.south);
      \draw[->] (sos.north) -- (decoder_rnn0.south);
      
      \draw[->] (decoder_fc0.north) -- (decoder0.south)node[label,midway,right] {$\vec{\hat{y}}^{(1)}$};
      \draw[->] (decoder_fc1.north) -- (decoder1.south)node[label,midway,right] {$\vec{\hat{y}}^{(2)}$};

      \draw[->] (decoder_rnn0.north) -- (decoder_fc0.south);
      \draw[->] (decoder_rnn1.north) -- (decoder_fc1.south);

      \coordinate[below left=0.75em and 0 of decoder_rnn0] (anchor_c0);
      \draw[-, draw=purple] (context.south) |- (anchor_c0);
      \draw[->, draw=purple] (anchor_c0) -| ([xshift=-0.5em]decoder_rnn0.south);

      \coordinate[below left=0.75em and 0 of decoder_rnn1] (anchor_c1);
      \draw[-, draw=purple] (anchor_c0) |- (anchor_c1);
      \draw[->, draw=purple] (anchor_c1) -| ([xshift=-0.5em]decoder_rnn1.south);

      \coordinate[below left=0.75em and 0 of decoder_rnn2] (anchor_c2);
      \draw[->, draw=purple] (anchor_c1) |- (anchor_c2);

      \coordinate[above right=0.5em and 0.5em of decoder0] (anchor_w10) {};
      \coordinate[below right=1.5em and 0.5em of decoder_rnn0] (anchor_w11) {};
      \draw[-] (decoder0.north) |- (anchor_w10);
      \draw[-] (anchor_w10) -- (anchor_w11);
      \draw[->] (anchor_w11) -| (decoder_rnn1.south);

      \coordinate[above right=0.5em and 0.5em of decoder1] (anchor_w20) {};
      \coordinate[below right=1.5em and 0.5em of decoder_rnn1] (anchor_w21) {};
      \draw[-] (decoder1.north) |- (anchor_w20);
      \draw[-] (anchor_w20) -- (anchor_w21);
      \draw[->] (anchor_w21) -| (decoder_rnn2.south);

      \draw[->,draw=green] (decoder_rnn0.east) -- (decoder_rnn1.west);
      \draw[->,draw=green] (decoder_rnn1.east) -- (decoder_rnn2.west);

      %
      %

      \node[encoder_rnn, left=3.5em of decoder_rnn0] (encoder_rnnN) {BRNNs};
      \node[encoder_rnn, left= of encoder_rnnN] (encoder_rnnt) {BRNNs};
      \node[encoder_rnn, left= of encoder_rnnt] (encoder_rnn1) {BRNNs};

      \node[embedding, below= of encoder_rnn1] (encoder_embedding1) {Embed};
      \node[embedding, below= of encoder_rnnt] (encoder_embeddingt) {Embed};
      \node[embedding, below= of encoder_rnnN] (encoder_embeddingN) {Embed};

      \node[input, below= of encoder_embeddingN] (mN) {$w^{(M)}$};
      \node[input, below= of encoder_embeddingt] (mt) {$\ldots$};
      \node[input, below= of encoder_embedding1] (m1) {$w^{(1)}$};

      \draw[->,draw=yellow] ([yshift=+0.5em]encoder_rnn1.east) -- ([yshift=+0.5em]encoder_rnnt.west);
      \draw[->,draw=yellow] ([yshift=+0.5em]encoder_rnnt.east) -- ([yshift=+0.5em]encoder_rnnN.west);
      \draw[<-,draw=yellow] ([yshift=-0.5em]encoder_rnn1.east) -- ([yshift=-0.5em]encoder_rnnt.west);
      \draw[<-,draw=yellow] ([yshift=-0.5em]encoder_rnnt.east) -- ([yshift=-0.5em]encoder_rnnN.west);
      \draw[->,draw=yellow] ([yshift=+0.5em]encoder_rnnN.east) -| (context.west);
      
      \coordinate[above left=0.5em and 0.5em of encoder_rnn1] (anchor_context_backward);
      \draw[-,draw=yellow] ([yshift=-0.5em]encoder_rnn1.west) -| (anchor_context_backward);
      \draw[->,draw=yellow] (anchor_context_backward) -| (context.north);

      \draw[->,draw=black] (m1.north) -- (encoder_embedding1.south);
      \draw[->,draw=black] (mt.north) -- (encoder_embeddingt.south);
      \draw[->,draw=black] (mN.north) -- (encoder_embeddingN.south);

      \draw[->,draw=blue] (encoder_embedding1.north) -- (encoder_rnn1.south);
      \draw[->,draw=blue] (encoder_embeddingt.north) -- (encoder_rnnt.south);
      \draw[->,draw=blue] (encoder_embeddingN.north) -- (encoder_rnnN.south);
  \end{tikzpicture}
}
        \caption{Language-to-motion model.}
        \label{fig:l2m}
    \end{subfigure}
    \caption{
        Overview of the proposed models for both directions.
        \autoref{fig:m2l} depicts the model that learns a mapping from a motion to language by first \emph{encoding} the motion sequence $\vec{M} = (\vec{m}^{(1)} \ldots \vec{m}^{(N)})$ into a \emph{context vector} $\vec{c}$~(in \textcolor{purple}{purple}) using a \emph{stack of bidirectional RNNs}~(BRNNs, in \textcolor{yellow}{yellow}).
        The context vector is then \emph{decoded} by \emph{another stack of unidirectional RNNs}~(in \textcolor{green}{green}), which also takes the \emph{embedded word}~(in \textcolor{blue}{blue}) generated in the previous timestep as input.
        A fully-connected layer~(FC, in \textcolor{lightgreen}{light green}) produces the parameters of the output probability distribution, denoted as $\vec{\hat{y}}^{(t)}$.
        The \emph{decoder}~(in \textcolor{red}{red}) finally takes $\vec{\hat{y}}^{(t)}$ and transforms it into a concrete word $\hat{w}^{(t)}$.
        Combined, the model thus generates the corresponding description word by word until the special \texttt{EOS} token is emitted and the description $\vec{\hat{w}} = (\hat{w}^{(1)}, \hat{w}^{(2)}, \ldots)$ is obtained.
        The other direction from language to motion is depicted in \autoref{fig:l2m}.
        The model works in similar fashion but uses a description in natural language $\vec{w} = (w^{(1)} \ldots w^{(M)})$ to generate the corresponding whole-body motion $\vec{\hat{M}} = (\vec{\hat{m}}^{(1)}, \vec{\hat{m}}^{(2)}, \ldots)$.
        The models for both directions are trained individually and share no weights.
    }
    \label{fig:models}
\end{figure}
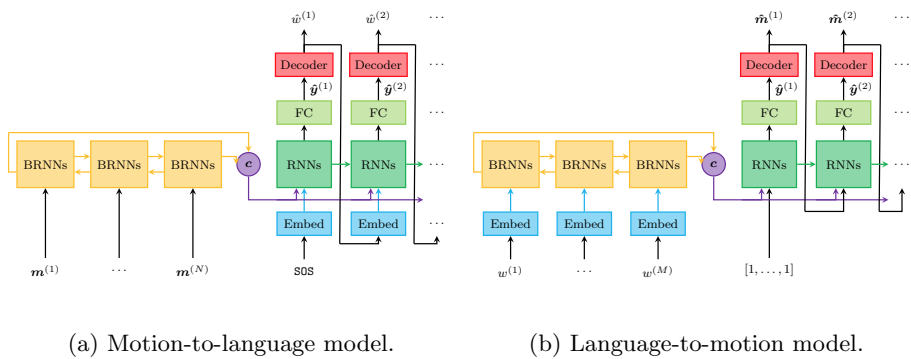

\subsection{Shared mode of operation}
Sequence-to-sequence learning has been used with great success, e.g. in large-scale machine translation~\citep{DBLP:journals/corr/WuSCLNMKCGMKSJL16}.
As the name suggests, the goal of such models is to generate a target sequence from an input sequence, where the sequences can differ in length and modality.
This property makes sequence-to-sequence learning an excellent fit for the purpose of learning a mapping between human motion and natural language.

We model each direction, that is from human motion to natural language as well as from natural language to human motion, individually.
\autoref{fig:models} depicts the details of both models.
In both cases, the input sequence is first transformed into a latent \emph{context vector} $\vec{c}$ by a recurrent neural network (RNN) or a stack thereof (meaning that many recurrent layers are stacked).
The output of the recurrent encoder network after the last timestep of the input sequence has been processed is used as the context vector. This context vector is then decoded by the \emph{decoder network} and, therefore, acts as the coupling mechanism between encoder and decoder network.
Different approaches have been proposed for this (e.g. initializing the hidden state of the decoder with the context vector) but we provide the context vector as input to the decoder network at each timestep.
The decoder then produces the desired target sequence step by step.

Typically, more advanced architectures than a vanilla RNN like \emph{long short-term memory}~(LSTM) \citep{lstm1, lstm2} or \emph{gated recurrent units}~(GRUs) \citep{gru1, gru2} are used.
The encoder often uses \emph{bidirectional RNNs}~(BRNNs)~\citep{brnn}, which process the input sequence in both directions and then combine the computed latent representations, e.g. by concatenation.
While we use GRUs and a bidirectional encoder in this work, the proposed model can be used with any recurrent network architecture.

We model the decoder output probabilistically, which means that the network predicts the parameters of some probability distribution instead of predicting the output value directly.
This intermediate output is denoted as $\vec{\hat{y}}^{(t)}$ and its form depends on the specific mapping (i.e. motion-to-language vs. language-to-motion) and are therefore described in \autoref{sec:m2l} and \autoref{sec:l2m} in more detail.

The \emph{decoder}\footnote{Note that the decoder is an element of the larger decoder network and the two are \emph{not} the same} finally decodes this probabilistic representation to a concrete deterministic instance.
One way to do this would be to greedily select the instance with highest probability under the distribution.
However, such a strategy does not necessarily yield the sequence with highest probability.
On the other hand, expanding each possible node is computational expensive for the discrete case and intractable for the continuous case.
We therefore use a common middle-ground between these two extremes: \emph{beam search}~\citep[chapter~12]{huang2001spoken}.
Beam search is a modification of best-first search, where in each step, only a limited set of stored nodes is considered for expansion.
Since the output of the network should depend on the decision of the decoder (which is made \emph{outside} of the decoder RNNs), we feed back this decision in each timestep.
The input at timestep $t$ is thus the concatenation of the decoded output from the previous timestep and the context vector, which is constant for all timesteps.

Finally, the encoder and decoder network can jointly be trained end-to-end using \emph{back-propagation through time}~(BPTT)~\citep{bptt}.

\subsection{Motion-to-language mapping}
\label{sec:m2l}
Having described the general sequence-to-sequence framework that is used throughout this work (depicted in \autoref{fig:m2l}), we now explain the concrete case of mapping from human motion to natural language.
To this end, the encoder network takes a motion sequence $\vec{M}$ as its input and encodes it into a context vector $\vec{c}$.
The decoder network then, step by step, produces the desired description in natural language $\vec{\hat{w}}~=~(\hat{w}^{(1)},~\hat{w}^{(2)},~\ldots)$ from this context vector, as described in the previous section.

\subsubsection{Model architecture}
The architecture of the encoder network is straightforward.
We use stacked bidirectional RNNs to compute the context vector given the motion sequence.
More concretely, we set the context vector to be the output of the last RNN layer after it has processed all timesteps of the input sequence.
In all layers, the outputs of the forward and backward processing (due to the bidirectional model) are concatenated before being passed to the next layer as input.

As mentioned before, our approach uses a probabilistic formulation of the decoder.
Fortunately, this can be achieved by defining a discrete probability distribution over the entire vocabulary.
This is realized in our model by a softmax layer as the final layer of the decoder network:
\begin{equation}
    \hat{y}^{(t)}_i = \frac{\exp{z_i^{(t)}}}{\sum_j{\exp{z_j^{(t)}}}},
\end{equation}
where $z_i^{(t)}$ denotes the unnormalized activation of the $i$-th output neuron, which corresponds to the $i$-th item in the vocabulary.
This can be interpreted as the probability of the $i$-th item in the vocabulary conditioned on the input motion encoded by the context vector and on the previously emitted words encoded by the hidden state of the recurrent decoder network:
\begin{equation}
    P\left(\hat{w}^{(t)}=i \mathrel{\Big|} \vec{M}, \hat{w}^{(t-1)}, \ldots, \hat{w}^{(1)}\right) := \hat{y}_i^{(t)}.
\end{equation}

The decoder network also uses stacked RNNs but connects them in such a way that each RNN has access to the context vector, the embedding of the previously emitted word and the output of the preceding RNN (if applicable).
Finally, the fully-connected softmax layer that produces the discrete probability distribution over the vocabulary as described above has access to the output of each RNN layer.
Details on the number of layers, units per layer and other hyperparameters are detailed in \autoref{sec:evaluation}.
A detailed schematic of the model architecture can also be found in \autoref{appendix:m2l_model}.

\subsubsection{Training}
We can train the entire model end-to-end by minimizing the categorical cross-entropy loss:
\begin{equation}
    \mathcal{L} = -\frac{1}{T} \sum_{t=1}^T{\sum_i{ y_i^{(t)} \log{\hat{y}_i^{(t)}} }}.
\end{equation}
$\vec{y}^{(t)}$ denotes the ground truth at timestep $t$ in the form of a reference natural language description and $\vec{\hat{y}}^{(t)}$ denotes the corresponding prediction of the model.
\added{Note that the loss is only computed for the active part of the description and does not include the padded part.}
We use BPTT with mini-batches to update the network parameters.
The exact hyperparameters of the training procedure are described in \autoref{sec:evaluation}.

\subsubsection{Decoding}
\label{sec:l2m-decoding}
During prediction, we face the problem of deciding on a concrete word $\hat{w}^{(t)}$ given the predicted probabilities $\vec{\hat{y}}^{(t)}$.
As mentioned before, we use beam search as a middle ground between greedily selecting the word with highest probability and performing an exhaustive search in the space of possible word sequences.
More concretely, we expand each of the candidate sequences by predicting $W$ different probability vectors $\vec{\hat{y}}_1^{(t)}$,~\ldots,~$\vec{\hat{y}}_K^{(t)}$ (beam search of width $W$) for each timestep.
Given a vocabulary of size $V$, this yields $W \cdot V$ new candidates, of which we only keep around the $W$ most likely sequences under our model.
This process is then repeated iteratively until each candidate has been completed as indicated by the \texttt{EOS} token, resulting in $W$ different descriptions.

Importantly, we also obtain the probability of each sequence by accumulating the product of all corresponding step-wise probabilities.
This in turn allows us to rank the candidates according to their probability under the model.

\subsection{Language-to-motion mapping}
\label{sec:l2m}
Although the approach for the language-to-motion mapping is similar to the previously describe motion-to-language mapping, the architecture of the decoder network is necessarily different since the output modality is now multi-dimensional and continuous.

\subsubsection{Model architecture}
The architecture of the encoder network is similar to what was already described in \autoref{sec:m2l}.
A network of stacked bidirectional RNNs encode the input description into a context vector.
The decoder uses stacked RNNs where each RNN layer has access to the context vector computed by the encoder network, the previously generated motion timestep and the output of the preceding RNN layers.
The complete model is depicted in \autoref{fig:l2m}.

However, there is a significant difference in the final fully-connected layer since the problem now requires to output a multi-dimensional and continuous frame of the motion~$\vec{\hat{m}}^{(t)}$ in each timestep (compared to the discrete word~$\hat{w}^{(t)}$ as in \autoref{sec:m2l}).
Furthermore, our approach uses a probabilistic decoder network, which allows us to generate non-deterministic outputs and also provides likelihood scores under our model for each generated sequence.

Recall that each motion frame is defined as ${\vec{\hat{m}}^{(t)} \in \mathbb{R}^{J} \times \{0, 1\}}$, where the first $J$~dimensions are the joint values (and therefore continuous) and the last dimension is the binary flag that indicates the active parts of the motion.
We use an approach similar to \cite{graves_rnns} based on a mixture of Gaussians for the continuous part and a Bernoulli distribution for the discrete part.

More concretely, the final fully-connected layer produces the following parameters, assuming a mixture model with $K$ components and omitting time indices for better readability:
\begin{itemize}
    \item $K$ component weights $\hat{\alpha}_1, \ldots, \hat{\alpha}_K \in [0, 1]$ produced by a softmax activation, and therefore $\sum_k{\hat{\alpha}_k} = 1$).
    \item $K$ mean vectors $\vec{\hat{\mu}}_1, \ldots, \vec{\hat{\mu}}_K \in \mathbb{R}^{J}$ produced by a linear activation.
    \item $K$ variance vectors $\vec{\hat{\sigma}}_1, \ldots, \vec{\hat{\sigma}}_K \in [0, \infty)^{J}$, assuming Gaussians with diagonal covariance matrices, produced by a softplus activation.
    \item The probability of being active ${\hat{p} \in [0, 1]}$ produced by a sigmoid activation.
\end{itemize}
To summarize, at each timestep the network predicts $$\vec{\hat{y}}^{(t)} = [(\alpha_i^{(t)}, \vec{\hat{\mu}}_i^{(t)}, \vec{\hat{\sigma}}_i^{(t)})_{i=1,\ldots,K}, \hat{p}^{(t)}],$$ which are the parameters for the mixture of Gaussians and Bernoulli distributions.

Given this formulation, we can define the likelihood of a given motion frame $\vec{\hat{m}}^{(t)}$ conditioned on the input description, as encoded by the context vector, as well as all previously emitted frames, encoded by the hidden state of the recurrent decoder network, as:
\begin{equation}
\begin{split}
  p\left(\vec{\hat{m}}^{(t)} \mathrel{\Big|} \vec{w}, \vec{\hat{m}}^{(1)}, \ldots, \vec{\hat{m}}^{(t-1)}\right) = \\
  \sum_{k=1}^K{\left[ \hat{\alpha}_k \ \mathcal{N}\left(\vec{\hat{m}}_{1:J}^{(t)} \mathrel{\Big|} \vec{\hat{\mu}}_k^{(t)}, \vec{\hat{\sigma}}_k^{(t)}\right)\right]} \cdot \mathcal{B}\left(\hat{m}_{J+1}^{(t)} \mathrel{\Big|} \hat{p}^{(t)} \right).
\end{split}
\end{equation}
Since we use diagonalized Gaussians, we can write the likelihood under each multivariate Gaussian as the product of $J$ one-dimensional Gaussians, one for each joint:
\begin{equation}
    \mathcal{N}\left(\vec{x} \mathrel{\Big|} \vec{\mu}, \vec{\sigma}\right) = \prod_{j=1}^{J}{\frac{1}{\sqrt{2 \pi \sigma_j^2}} \exp{\left( -\frac{(x_j - \mu_j)^2}{2 \sigma_j^2} \right)}}.
    \label{eq:l2m_prod}
\end{equation}
Finally, the probability mass for the discrete Bernoulli distribution is defined as
\begin{equation}
  \mathcal{B}\left(x \mathrel{\Big|} p \right) =
  \left\{
  \begin{array}{@{}ll@{}}
    p, & \text{if} \ x = 1 \\
    1-p, & \text{otherwise.}
  \end{array}\right.
\end{equation}

Details on the number of layers, units per layer and other hyperparameters are detailed in \autoref{sec:evaluation}.
A detailed schematic of the model architecture can also be found in \autoref{appendix:l2m_model}.

\subsubsection{Training}
Again, we can train the entire model end-to-end by minimizing the following loss:
\begin{equation}
\begin{split}
    \mathcal{L} = \frac{1}{T} \sum_{t=1}^T \left[
        -\log \sum_k{\hat{\alpha}_k \ \mathcal{N}\left( \vec{m}_{1:J}^{(t)} \mathrel{\Big|} \vec{\hat{\mu}}^{(t)}, \vec{\hat{\sigma}}^{(t)} \right)} \right. \\
        \left. - m_{J+1}^{(t)} \log{\hat{p}^{(t)}} - \left(1 - m_{J+1}^{(t)}\right) \log{\left(1 - \hat{p}^{(t)}\right)}
     \right].
\end{split}
\label{eq:l2m_loss}
\end{equation}
\added{Note that the loss is only computed for the active part of the motion and does not include the padded part.}

This loss consists of two parts.
The first part describes the likelihood of the ground truth joint values (denoted as $\vec{m}_{1:J}^{(t)}$) under the predicted mixture distribution.
The second part is the binary cross-entropy between the ground truth of the active flag $m_{J+1}^{(t)}$ and the predicted parameter of the Bernoulli distribution $\hat{p}^{(t)}$.
By minimizing the loss $\mathcal{L}$, we jointly maximize the likelihood of the ground truth under the mixture of Gaussians and minimize the cross-entropy.

However, in practice the formulation in \autoref{eq:l2m_loss} has numerical stability issues.
This is because computing the likelihood under each multi-variate Gaussian requires computing the product of $J$~individual likelihoods (compare \autoref{eq:l2m_prod}), which can easily be subject to both numerical under- and overflow.

We therefore define the following surrogate loss function:
\begin{equation}
\begin{split}
    \mathcal{\widetilde{L}} = \frac{1}{T} \sum_{t=1}^T \left[
        -\sum_k{\hat{\alpha}_k^{(t)} \ \log \mathcal{N}\left( \vec{m}_{1:J}^{(t)} \mathrel{\Big|} \vec{\hat{\mu}}^{(t)}, \vec{\hat{\sigma}}^{(t)} \right)} \right. \\
        \left. - m_{J+1}^{(t)} \log{\hat{p}^{(t)}} - \left(1 - m_{J+1}^{(t)}\right) \log{\left(1 - \hat{p}^{(t)}\right)} \right]
\end{split}
\end{equation}
This re-formulation allows us to replace the product from \autoref{eq:l2m_prod} with a sum of logarithms, making the computation more robust against numerical problems.
We find that minimizing $\mathcal{\widetilde{L}}$ instead of $\mathcal{L}$ works well and yields the desired results.

Like before, we use BPTT with mini-batches to train the network end-to-end.
Again, the exact hyperparameters of the training procedure are described in \autoref{sec:evaluation}.

\subsubsection{Decoding}
Similar to motion-to-language mapping (\autoref{sec:l2m-decoding}), we again face the problem of decoding a concrete motion frame $\vec{\hat{m}}^{(t)}$ from the prediction vector $\vec{\hat{y}}^{(t)} = [(\alpha_i^{(t)}, \vec{\hat{\mu}}_i^{(t)}, \vec{\hat{\sigma}}_i^{(t)})_{i=1,\ldots,K}, \hat{p}^{(t)}]$.
This can be solved by sampling the joint values from the multivariate Gaussian mixture distribution (by first sampling a mixture component with probabilities $\alpha_1^{(t)}, \ldots, \alpha_K^{(t)}$ and then sampling from the selected multivariate Gaussian) as well as sampling from the Bernoulli distribution for the binary active indicator.
Like before we use beam search to obtain a couple of candidates at the end of the decoding process as previously described in \autoref{sec:l2m-decoding}.
The key difference here is that we cannot compute the likelihood for each discrete possibility anymore.
We resolve this problem by sampling a couple of candidates for each hypothesis from the respective distributions and then truncating them to keep around a fixed number of hypotheses for the next timestep.

\section{Experiments}
\label{sec:evaluation}
\begin{sidewaystable}
    \caption{The hyperparameters for both models that were used for all experiments. When using bidirectional RNNs, the number of units per layer is given in the form $2 \times x$, where $x$ is the number of units for each processing direction.}
    \label{table:model-hyperparams}
    \tvspace{1.12}
    \thspace{0.5cm}
    \begin{center}
        \begin{tabular}{@{}lrr@{}} \toprule
            \textbf{Hyperparameter}        & \textbf{Motion-to-Language}  & \textbf{Language-to-Motion}  \\ \midrule
            Encoder RNN type               & bidirectional GRU            & bidirectional GRU with layer normalization  \\
            Encoder layers                 & $2$ layers $(2 \times 64, 2 \times 64)$      & $2$ layers $(2 \times 64, 2 \times 64)$       \\ 
            Decoder RNN type               & GRU                          & GRU with layer normalization  \\
            Decoder layers                 & $2$ layers $(128, 128)$      & $3$ layers $(400, 400, 400)$  \\
            Embedding dimension            & $64$                         & $64$                          \\
            Dropout rate                   & $0.4$                        & $0.1$                         \\
            Optimizer                      & Adam with Nesterov           & Adam with Nesterov            \\
            Learning rate                  & $10^{-3}$                    & $10^{-3}$                     \\
            Gradient clipping              & $\infty$                     & $25$                          \\
            Batch size                     & $128$                        & $128$                         \\
            Training epochs                & $100$                        & $100$                         \\
            Vocabulary size ($V$)          & $1\,344$                     & $1\,344$                      \\
            Motion joints ($J$)            & $44$                         & $44$                          \\
            Mixture components             & \emph{n/a}                   & $20$                          \\ \midrule
            Trainable parameters           & $843\,456$                   & $5\,446\,517$                 \\
            \quad excl. embedding \& FC layers & $414\,720$                   & $3\,221\,520$                 \\
            \bottomrule
        \end{tabular}
    \end{center}
\end{sidewaystable}

\subsection{Dataset}
While large datasets for human motion exist (see also \cite{DBLP:journals/trob/ManderyTDVA16} for a recent review), our evaluation requires a dataset that also contains descriptions of such motion in natural language.
We have recently proposed the \emph{KIT Motion-Language Dataset}~\citep{DBLP:journals/corr/PlappertMA16}, which uses human whole-body motions from the \emph{KIT Whole-Body Human Motion Database}\footnote{\url{https://motion-database.humanoids.kit.edu/}}~\citep{DBLP:conf/icar/ManderyTDVA15} and the \emph{CMU Graphics Lab Motion Capture Database}\footnote{\url{http://mocap.cs.cmu.edu/}}.
For each motion, a set of descriptions in the form of a single English sentence was collected using a crowd-sourcing approach.

We use the \texttt{2016-10-10} release of the KIT Motion-Language Dataset for all experiments.
This version of the dataset contains $3\,911$~recordings of human whole-body motion in the aforementioned MMM representation and $6\,278$~annotations in natural language.
The dataset is publicly available\footnote{\url{https://motion-annotation.humanoids.kit.edu/dataset/}} so that all results in this paper can be reproduced.

For our evaluation, we filter out motions that have a duration of $30$~seconds or more in order to reduce the computational overhead due to excessive padding.
\added{We have decided to discard these motions for two reasons.
First, almost all motions are below $30$ seconds in duration, which means that almost no data is discarded to begin with.
Second, we decided to discard motions instead of clipping them since we otherwise cannot guarantee that all important parts are visible to our model, i.e. we may remove crucial information that characterizes a motion.}

This results in $2\,846$~usable motion samples with a total duration of $5.3$~hours and $6\,187$~natural language annotations that consist of $46\,561$~words in total, with a vocabulary size of $1\,344$.
We randomly split the remaining data into training, validation and test sets with a ratio of $0.8$, $0.1$ and $0.1$, respectively.
\secondreview{We compute the split such that a motion and all its associated natural language descriptions are always in the same set and are not partially leaked into another set.}
All results are reported using the test set, if not otherwise indicated.
The processing steps described in \autoref{sec:representation} are performed to obtain the representations suitable for training the model.

\subsection{Setup}
The model architectures for motion-to-language (\autoref{sec:m2l}) and language-to-motion (\autoref{sec:l2m}) have already been described.
However, we have not yet described the specific hyperparameters that we used for our experimental results.

For both the encoder and decoder parts of each model, we use \emph{gated recurrent units} (GRUs)~\citep{gru1} as the recurrent neural network architecture.
We regularize all models with \emph{dropout}~\citep{dropout, dropout_rnn} in the embedding, recurrent and fully-connected layers.
In both cases, we train our models using the \emph{Adam} optimizer with Nesterov momentum~\citep{adam, nadam}.
Training of the language-to-motion model proved to be more difficult, presumably due to the more complex dynamics of the recurrent model, which made it necessary to use both \emph{gradient clipping}~\citep{gradient_clipping} as well as \emph{layer normalization}~\citep{layernorm}.
We also experimented with \emph{batch normalization}~\citep{batchnorm, batchnorm_rnn} instead of layer normalization but were unable to get good results with it, presumably due to the known problems of batch normalization with padded sequences.\footnote{\url{https://github.com/cooijmanstim/recurrent-batch-normalization/issues/2}}

\autoref{table:model-hyperparams} summaries the aforementioned and lists all other hyperparameters for both models, language-to-motion and motion-to-language.
These hyperparameters were used for all experiments throughout this work.
The code that was used to obtain all following results is available online: \url{https://gitlab.com/h2t/DeepMotionLanguageMapping/}

\subsection{Generating language from motion}
\subsubsection{Training}
\begin{figure}[h]
    \centering
    \includegraphics[width=0.7\textwidth]{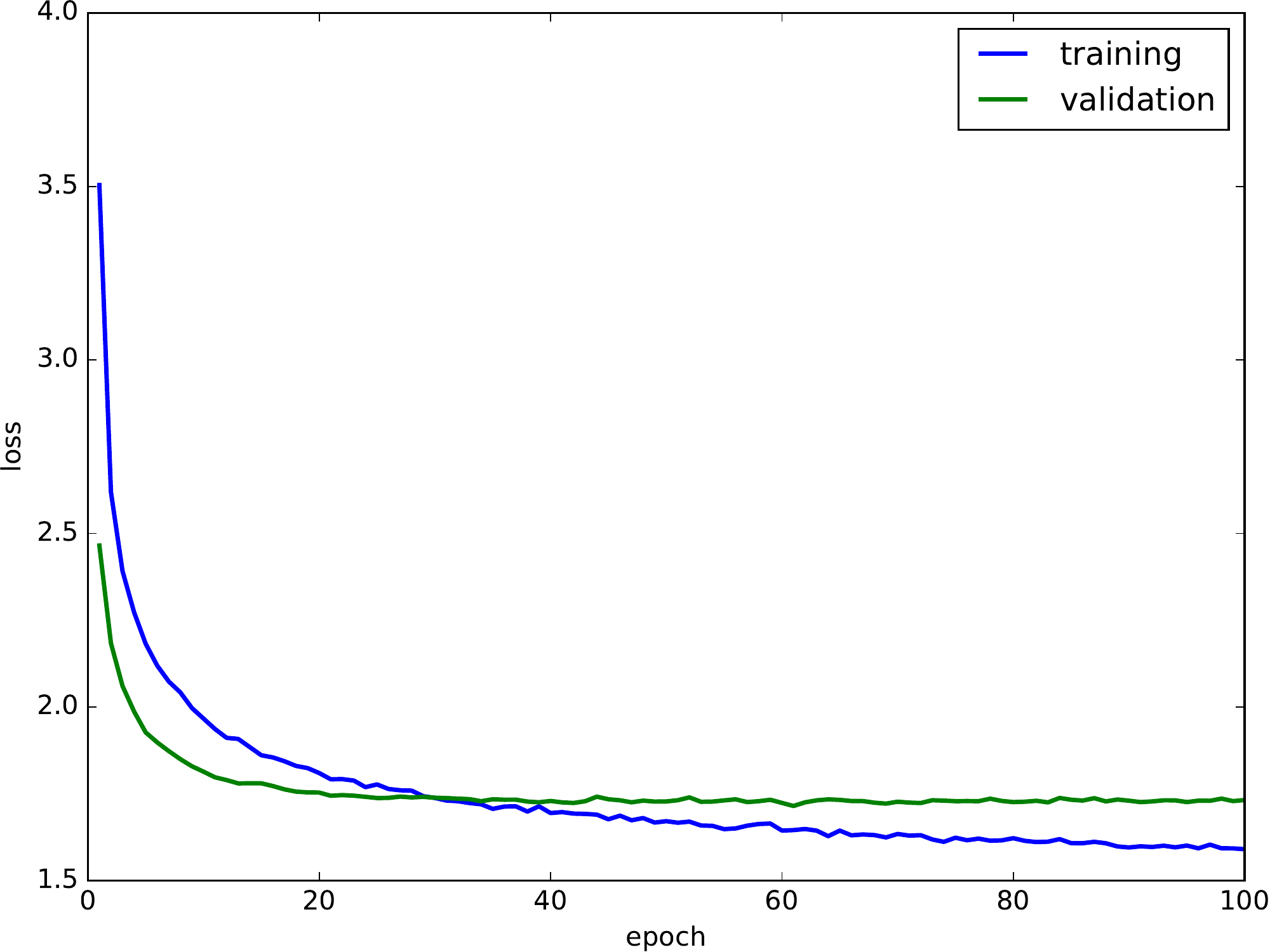}
    \caption{Training and validation loss during training of the motion-to-language model.}
    \label{fig:m2l_loss}
\end{figure}

Optimizing the model turned out to be straightforward and did not require special measures.
\autoref{fig:m2l_loss} depicts the change in loss during training for the training and validation split, respectively.
As can be seen, both training and validation loss continuously decrease, indicating that the model does not overfit on the training data.
Additionally, the validation loss seems to have converged after $100$~training epochs, indicating that our training time was sufficiently long.
\added{Training took approximately 5 hours on an Nvidia GeForce GTX 980 Ti graphics card with an Intel Core i7-6700K.
After training, our model can generate approximately 12 descriptions per second.}

\subsubsection{Qualitative results}
\begin{figure}[b]
    \centering
    \scalebox{0.9}{
        \begin{minipage}{\textwidth}
            \input{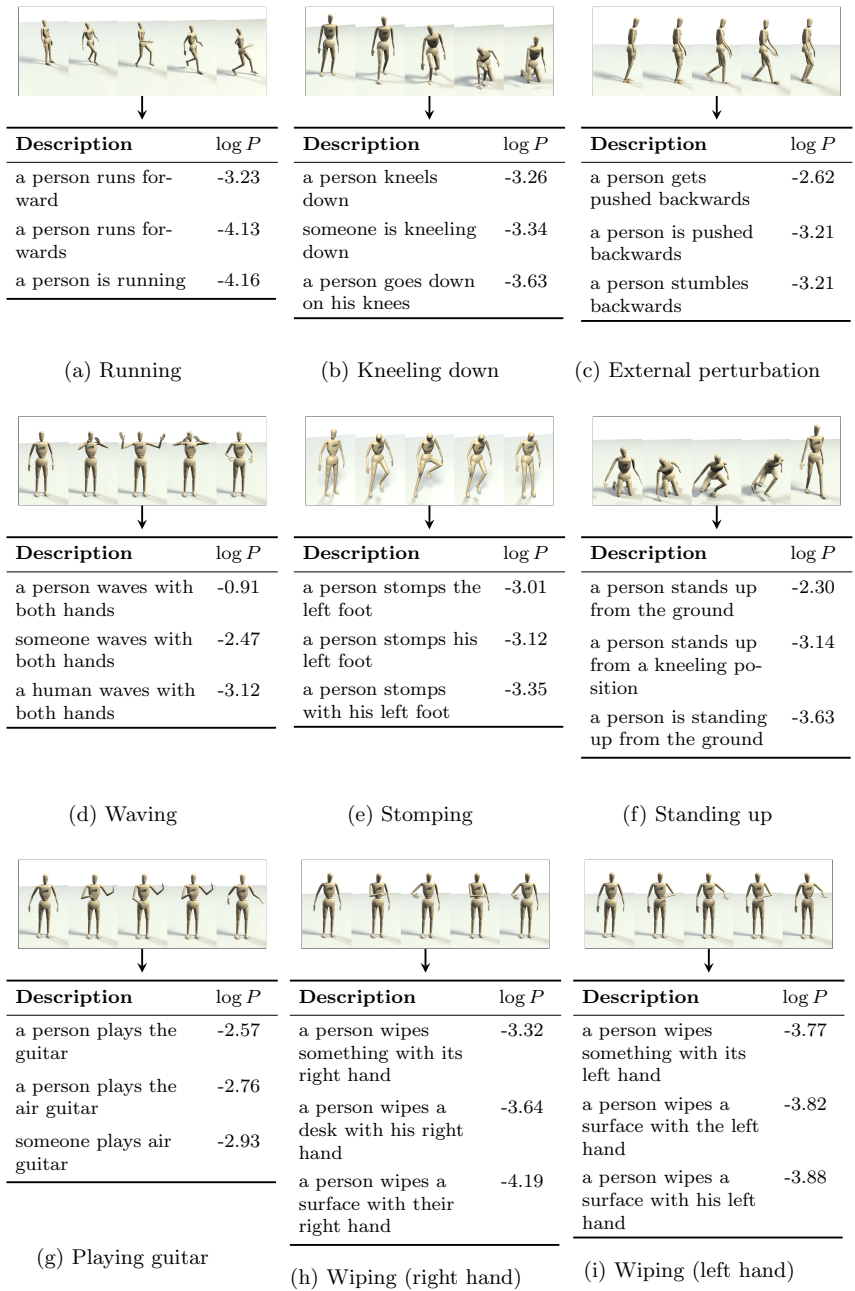}
        \end{minipage}
    }
    \caption{Exemplary natural language descriptions as generated by the proposed model for nine different human whole-body motions.
    The description hypotheses are sorted in descending order by their log probability under the model.}
    \label{fig:m2l_examples}
\end{figure}

To provide insight into the style and quality of the natural language descriptions generated by our model,
we present a few examples in \autoref{fig:m2l_examples}.
For each depicted human whole-body motion, we compute five natural language description hypotheses (thus, we perform beam search with ${W=5}$) and sort them in descending order by their respective log probability under the model (depicted in the tables below each motion; due to space constraints, we only present the top three descriptions per motion).
As can be seen from these examples, all descriptions are complete English sentences with valid grammatical structure.
This is a pattern that we observe throughout.
Interestingly, the model also produces semantically identical descriptions with varying grammatical structure.
For example, when describing standing up~(\autoref{fig:m2l_examples_stand_up}), the model creates a description using present simple~(\emph{``a person stands up from the ground''}) and another one using present continuous~(\emph{``a person is standing up from the ground''}).

The model also uses synonyms interchangeably.
For example, the model refers to the subject in the scene as a \emph{``human''}, \emph{``person''} or \emph{``someone''}.
This is especially apparent in \autoref{fig:m2l_examples_wave}, where the produced sentences are identical except for this variation.

The generated natural language descriptions are also rich in detail.
For example, the model successfully differentiates between wiping a surface with the right~(\autoref{fig:m2l_examples_wipe_right}) vs. left~(\autoref{fig:m2l_examples_wipe_left}) hand and creates corresponding descriptions that mention the handedness of the motion.
Similar behavior can be observed for the waving motion~(\autoref{fig:m2l_examples_wave}), stomping motion~(\autoref{fig:m2l_examples_stomp}) and pushing motion~(\autoref{fig:m2l_examples_push}), for which the descriptions all correctly mention the correct hand, foot and perturbation direction, respectively.

Overall, the contents of the generated descriptions are highly encouraging and demonstrate the capabilities of the model to not only generate syntactically valid descriptions, but also semantically meaningful and detailed ones.

\FloatBarrier

\subsubsection{Quantitative results}
\label{sec:m2l-quantitative}
While the previously presented results allow us to gather some understanding of the generated descriptions, they do not necessarily represent the overall behavior of the model.
Therefore, we provide in the following a quantitative evaluation of the performance of the proposed model over the training and test splits of our dataset.

We select the \textsc{Bleu} score~\citep{DBLP:conf/acl/PapineniRWZ02} to measure the performance of the proposed model.
Briefly speaking, the \textsc{Bleu} score is a metric that was initially proposed to measure the performance of machine translation systems and has also been used in work targeting the similar problem of learning a mapping between human motion and natural language~\citep{DBLP:journals/nn/TakanoKN16}.

The \textsc{Bleu} score is obtained by first counting unigrams, bigrams, \ldots, $n$-grams (in this work up to $n=4$) in the provided reference and then computing the $n$-gram precision (with exhaustion) of the hypothesis.
Essentially, it is this modified and weighted $n$-gram precision with some additional smoothing.
The \textsc{Bleu} score is defined to be between $0$ and $1$, with $1$ indicating a perfect match, i.e. the hypothesis is word for word identical to one of the references.

We compute the \textsc{Bleu} score on a corpus-level, in contrast to sentence-level as in other works on this problem.
This is an important distinction, since sentence-level computation estimates the $n$-gram model only on very few reference sentences, leading to bad estimates and therefore unreliable scores.
\added{In contrast, for corpus-level \textsc{Bleu} scores, all provided descriptions are used to compute the probabilities of the $n$-gram model.}
More concretely, we generate five descriptions, that is the hypotheses, for each human motion and sort in descending order by their log probability under the model.
We then compute five different \textsc{Bleu} scores that correspond to selecting the 1st, 2nd, \ldots, 5th best hypothesis.
In all cases, we use all available annotations, which have been created by the human annotators, as the ground truth.
\textsc{Bleu} scores are computed separately for the training and test split.
\added{It should also be noted that the \textsc{Bleu} score is not without flaw when it comes to evaluations outside of the machine translation domain~\citep{DBLP:journals/jair/HodoshYH13}.
However, we opt to use it here since most prior work is evaluated using the \textsc{Bleu} score.}

\begin{table}[t]
    \caption{Corpus-level \textsc{Bleu}~scores for the motion-to-language model.}
    \label{table:m2l_bleu}
    \tvspace{1.12}
    \thspace{0.3cm}
    \begin{center}
        \begin{tabular}{@{}lrrrrr@{}} \toprule
                            & \multicolumn{5}{c}{\textsc{Bleu} \textbf{scores}} \\
                              \cmidrule{2-6}
                            & 1st     & 2nd      & 3rd     & 4th     & 5th  \\ \midrule
            \textbf{Train}  & 0.387   & 0.355    & 0.330   & 0.329   & 0.302    \\
            \textbf{Test}   & 0.338   & 0.283    & 0.295   & 0.277   & 0.250    \\
            \bottomrule
        \end{tabular}
    \end{center}
\end{table}

\autoref{table:m2l_bleu} lists the achieved \textsc{Bleu} scores.
A couple of observations are noteworthy.
First, the \textsc{Bleu} scores are lower than one would expect from a machine translation system, for example.
This is because there is much more ambiguity when generating descriptions of human motion than when translating a text into a different language.
In the former case, different levels of details and different styles are also semantically correct (e.g. \emph{``A human walks''} vs. \emph{``Someones takes a couple of steps''}), whereas the latter case is much more constrained.
Second, the \textsc{Bleu} scores are clearly correlated with the order of the hypotheses (as defined by their log probabilities under the model) even though the loss that was used to train the model and the \textsc{Bleu} score are completely separate.
This means that the log probability is suitable as a measure of quality and, in turn, that the model indeed captures some understanding of what an objectively high-quality (as measured by \textsc{Bleu}), description is.

Third, the model achieves slightly worse, but still comparable performance on the test split.
This, again, demonstrates that the model does not overfit on the training data and generalizes to previously unseen motions.
Additionally, the second point, that \textsc{Bleu} scores and their ranking as defined by the respective log probabilities of the hypotheses are correlated, still holds; however, the pattern is a bit more noisy.

\subsection{Generating motion from language}

\subsubsection{Training}
\begin{figure}[h]
    \centering
    \includegraphics[width=0.7\textwidth]{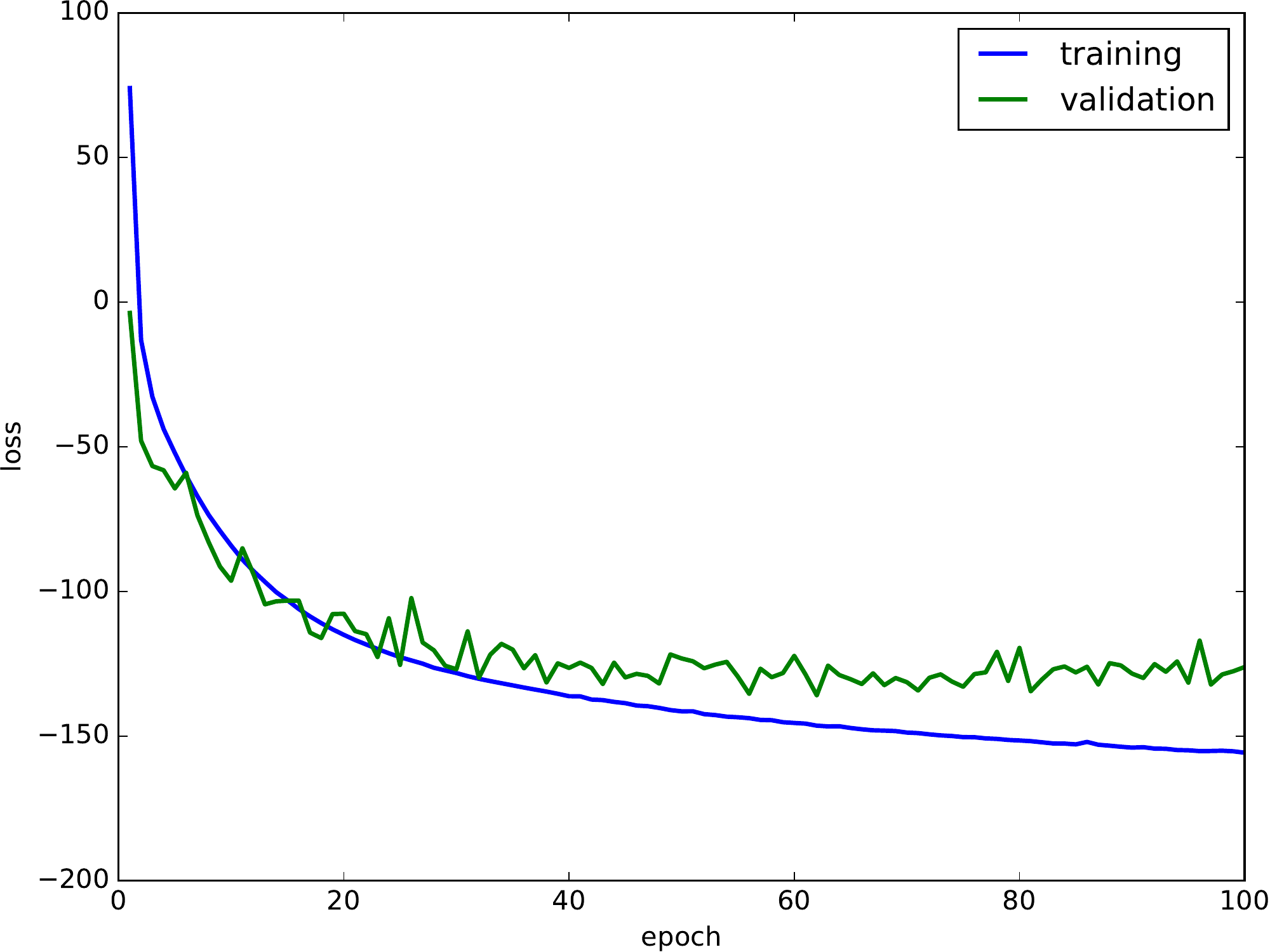}
    \caption{Training and validation loss during training of the language-to-motion model.}
    \label{fig:l2m_loss}
\end{figure}

Optimizing the language-to-motion model proved to be more complex, which was presumably due to the much more complex decoder network.
In order to successfully optimize the model, we found that gradient clipping and layer normalization play a crucial role.
The learning curves for the training and validation splits are depicted in \autoref{fig:l2m_loss}.
\added{Training took approximately 24 hours on an Nvidia GeForce GTX 980 Ti graphics card with an Intel Core i7-6700K.
After training, our model can generate approximately 5 motions per second.}

Although the curve for the validation loss is more noisy than in the motion-to-language case, the model still appears to learn the problem at hand without overfitting on the training data.

\subsubsection{Qualitative results}
\begin{figure}[h]
    \input{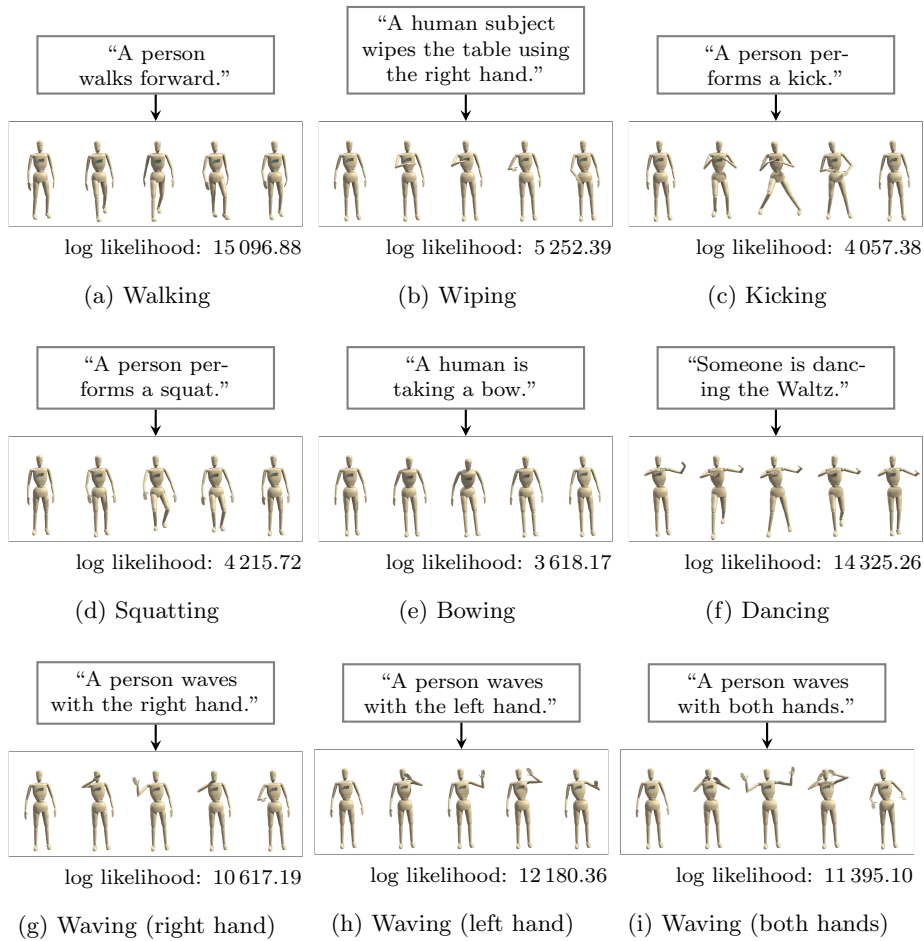}
    \caption{Exemplary human whole-body motions as generated by the proposed model from different natural language descriptions.}
    \label{fig:l2m_examples}
\end{figure}

Similar to before, we provide some insight into the type of motion the proposed model generates by visualizing several examples in~\autoref{fig:l2m_examples}.
For each given natural language description, we compute five human whole-body motion hypotheses (thus ${W=5}$).
Due to space constraints, we only depict the motion with the highest loglikelihood under the model.

Most interestingly, the results also demonstrate that our model is not only capable of generating the correct motion primitive, but also to adjust the generated motion to a desired parametrization, e.g. the model generates motions for waving with the right, left and both hands~(\autoref{fig:l2m_examples_wave_right}, \autoref{fig:l2m_examples_wave_left} and \autoref{fig:l2m_examples_wave_both}.
We observed similar behavior in other examples such as walking motions with different speed.

For periodic movements, we observed that the model does generate motions with a number of repetitions (e.g. waving seven times) that were not indeed included in the training set, suggesting that it does discover the underlying periodic structure of the motion.
However, we were unable to parametrize the number of repetitions using language.
We hypothesize that this is because the training data does not contain enough training examples to learn counting.

Another limitation of the proposed model is caused by the fact that we represent the motion using only joint angles.
This, in turn, means that the model does not predict the pose of the human subject in space, which can be observed for the bowing, squatting and kicking motions~(\autoref{fig:l2m_examples_bow}, \autoref{fig:l2m_examples_squat} and \autoref{fig:l2m_examples_kick}), where the root pose of the human model remains fixed.
Furthermore, since we only consider the kinematic aspects of a motion, the generated motions are not necessarily dynamically stable.
Similarly, since we do not consider contact information with objects at this point, generated manipulation motions are violating constraints that would be necessary to achieve the desired outcome.
This can be observed for the wiping motion~(\autoref{fig:l2m_examples_wipe}), where the wiping diverges from the (imaginary) table surface over time.
Including these dynamic properties and contact information is an important area of future work.

Overall, we clearly demonstrate the capabilities of the model to generate a wide variety of realistic human whole-body motions specified only by a description in natural language.
We also visualize the entirety of the generated motions in a supplementary video: \url{https://youtu.be/2UQWOZtsg-8}.

\subsubsection{Quantitative results}
Providing quantitative results for the performance of the language-to-motion model is complicated since defining an appropriate metric is non-trivial.
The reason for this is that a motion can be performed in a large variety of different styles that can all be semantically correct for the given description but may have very different joint-level characteristics.
Computing a metric like the mean squared error between reference and hypotheses is therefore ill-suited to judge how correct the motion hypothesis is \emph{semantically}.

\added{An obvious choice for evaluation would be a user study.
Unfortunately, this comes with a significant cost. Our dataset contains $2\,846$ motions and $6\,187$ descriptions in natural language.
If we generate 5 hypotheses per motion and description, we obtain $14\,230$ descriptions (for the $2\,846$ motions) and $24\,748$ motions (for the $6\,187$ descriptions).
To be representative, we would need multiple users to review each of these generated motions and descriptions, which would take a significant amount of time\secondreview{, even if we would only evaluate a subset}.
Additionally, the design of such a user study is quite non-trivial due to inherent ambiguous nature of the problem. 
For example, consider a waving motion in which the subject waves with the left hand.
It is unclear if a generated description that describes this handedness should be preferred over a description that does not describe it since both are equally valid and simply describe the same thing with varying levels of detail.
Additionally, we believe that a user study alone would be problematic since it makes it almost impossible for other authors to compare their systems with ours.}

\added{Another possible way of evaluating our model would be to train a classifier to predict the type of the ground truth motion data and then use this classifier to estimate the quality of the generated motion data.
The problem with this approach is that it would require labels for each motion.
Even if these labels would be available, a classifier with low prediction error would need to be trained, which is in itself a non-trivial task.
Even then, we would still face the problem of ambiguity.
Consider, for example, a description of a waving motion that does not specify the handedness.
If the generated motion does wave with the left hand, but the ground truth data was actually waving with the right hand, the motion would be counted as incorrect.
However, given the available data, the generator clearly successfully generated a motion and the description was simply not detailed enough.}

To resolve these problems, we exploit the fact that we already can compute a semantic description of a given motion using our previously evaluated motion-to-language model.
Since we have already evaluated this model separately, we can use its performance as a baseline to judge the quality of the other direction, namely language-to-motion.
\added{Additionally, this evaluation does not require a separate system like a classifier and scales to many thousands of motion-language tuples.}

More concretely, we essentially chain the two models.
First, we use the language-to-motion model, which we want to evaluate here, to compute a motion given a description in natural language.
Next, we use the previously trained and evaluated motion-to-language model to transform this generated motion back into a description in natural language.
Finally, we can compute the \textsc{Bleu}~score as described in~\autoref{sec:m2l-quantitative} to quantitatively measure the performance of the language-to-motion model.
The results of this approach are given in~\autoref{table:l2m_bleu}.

\begin{table}[t]
    \caption{Corpus-level \textsc{Bleu}~scores for the language-to-motion model.}
    \label{table:l2m_bleu}
    \tvspace{1.12}
    \thspace{0.3cm}
    \begin{center}
        \begin{tabular}{@{}lrrrrr@{}} \toprule
                            & \multicolumn{5}{c}{\textsc{Bleu} \textbf{scores}} \\
                              \cmidrule{2-6}
                            & 1st     & 2nd      & 3rd     & 4th     & 5th  \\ \midrule
            \textbf{Train}  & 0.256   & 0.277    & 0.289   & 0.286   & 0.278    \\ 
            \textbf{Test}   & 0.249   & 0.242    & 0.288   & 0.249   & 0.240    \\
            \bottomrule
        \end{tabular}
    \end{center}
\end{table}

To make it easier to compare the performance, we propose to measure the language-to-motion model relative to the performance of the motion-to-language model.
More formally, we relate the \textsc{Bleu} score of the language-to-motion model (to be evaluated, see \autoref{table:l2m_bleu}) to the highest \textsc{Bleu} score of the motion-to-language model (the baseline, which is $0.387$) by dividing the two.
We thus measure the percentage of performance that is retained after transforming the natural language into a motion hypothesis and then transforming this generated motion back into a description.
\autoref{table:l2m_bleu_rel} lists this relative performance for the language-to-motion model.

\begin{table}[b]
    \caption{Relative performance of the language-to-motion model compared to the baseline performance of the motion-to-language model.}
    \label{table:l2m_bleu_rel}
    \tvspace{1.12}
    \thspace{0.25cm}
    \begin{center}
        \begin{tabular}{@{}lrrrrr@{}} \toprule
                            & \multicolumn{5}{c}{\textbf{Relative Performance}} \\
                              \cmidrule{2-6}
                            & 1st     & 2nd      & 3rd     & 4th     & 5th  \\ \midrule
            \textbf{Train}  & 66.1\%   & 71.6\%    & 74.7\%   & 73.9\%   & 71.8\%    \\ 
            \textbf{Test}   & 64.3\%   & 62.5\%    & 74.4\%   & 64.3\%   & 62.0\%    \\
            \bottomrule
        \end{tabular}
    \end{center}
\end{table}

A couple of observations.
First, the results clearly demonstrate that our model is capable of generating the correct human motion beyond the few examples presented in the qualitative analysis described before.
Compared to the baseline, the model achieves a performance of $71.6\% \pm 3.0$ for both the training and $65.5\% \pm 4.5$ for the test split.
As expected, the \textsc{Bleu} scores for the training split are slightly higher than for the test split, but still comparable.
This suggests minimal overfit and generalization capabilities to previously unseen examples.
Second, the ranking of hypotheses as defined by their loglikelihood under the model seems to be less correlated with the performance of the model under the proposed evaluation metric.
Currently, it is difficult to identify the prime origin and cause for this since this could also be caused by the error of the motion-to-language model.

\subsection{Understanding the model}
Our final experiment is concerned with providing insight into the latent representations that the proposed model uses internally.
In order to do so, we analyze the context vectors produced by both the motion-to-language model and language-to-motion model.
Since they are extremely high-dimensional~(${\vec{c} \in \mathbb{R}^{128}}$), we use \emph{t-distributed Stochastic Neighbor Embedding}~(t-SNE)~\citep{tsne} to project all context vectors into a 2-dimensional representation.
t-SNE is particular well-suited for this task since it is known to maintain the structure of the original high-dimensional vector space in the low-dimensional projection.

\begin{figure}[h]
    \centering
    \begin{subfigure}[t]{0.47\textwidth}
        \includegraphics[width=\textwidth,trim=1.5cm 0.5cm 1.5cm 1cm,clip=true]{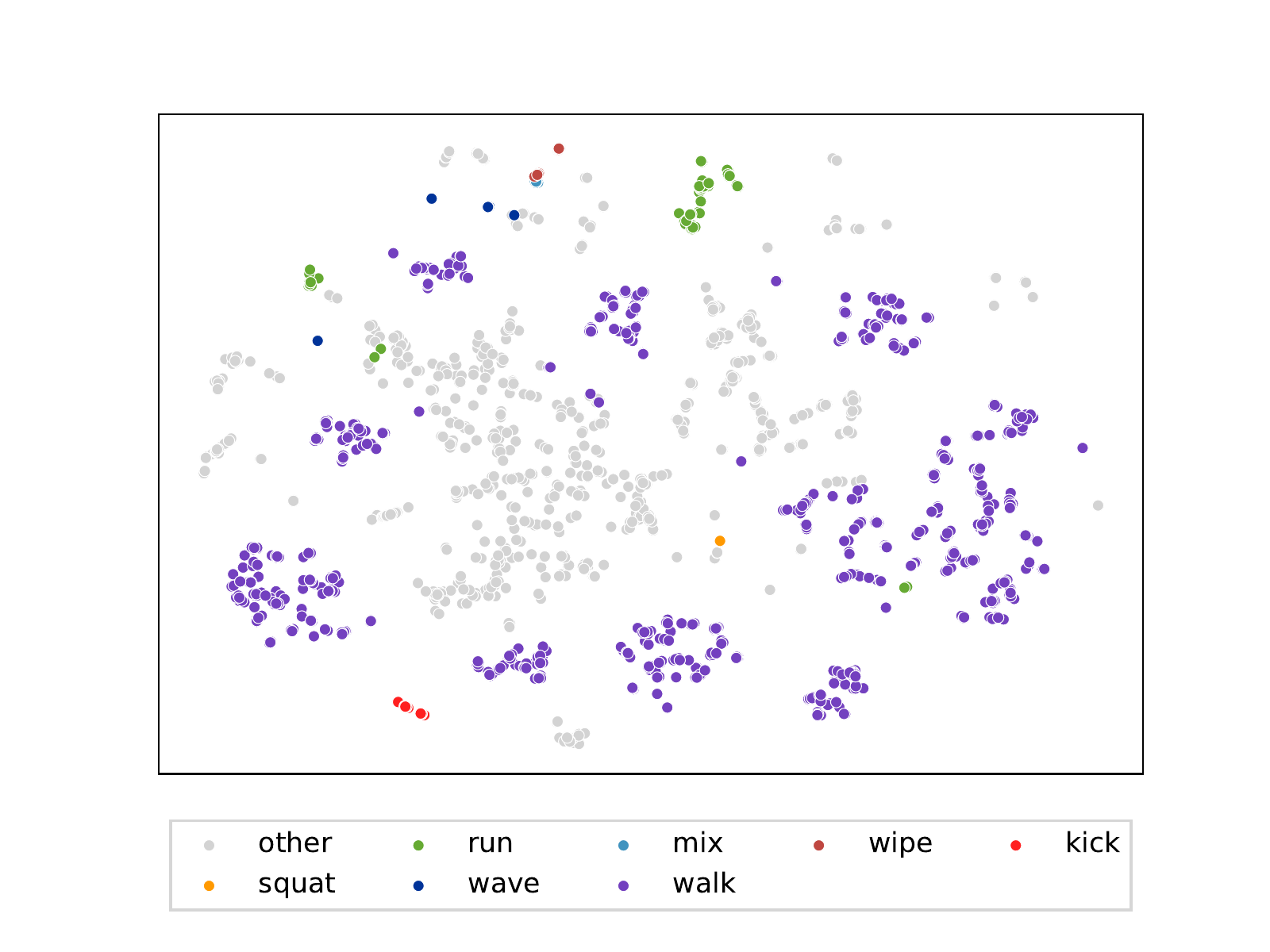}
        \caption{Motion-to-language context vectors.}
        \label{fig:m2l_contexts}
    \end{subfigure}
    \hfill
    \begin{subfigure}[t]{0.47\textwidth}
        \includegraphics[width=\textwidth,trim=1.5cm 0.5cm 1.5cm 1cm,clip=true]{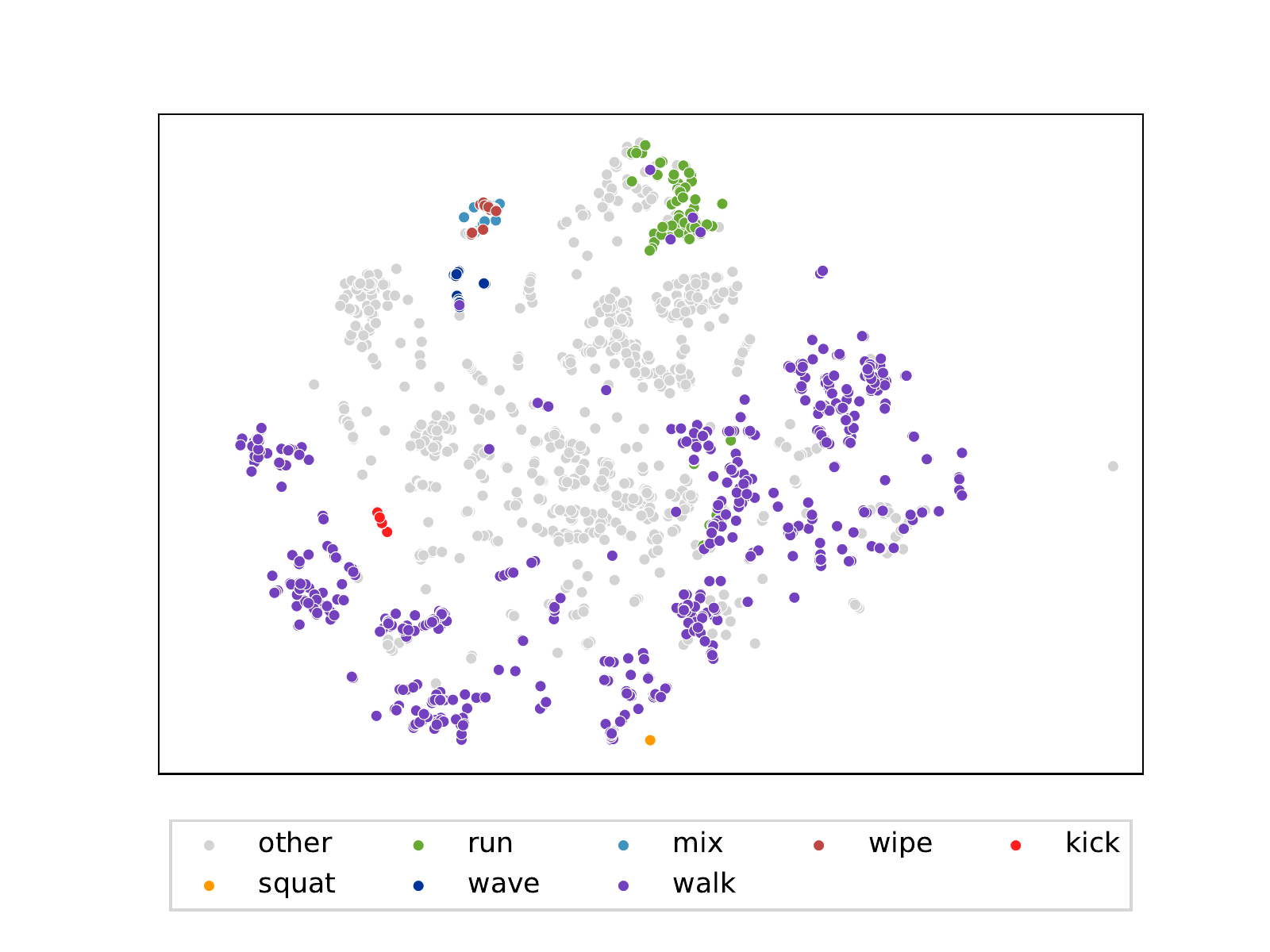}
        \caption{Language-to-motion context vectors.}
        \label{fig:l2m_contexts}
    \end{subfigure}
    \caption{Visualization of the context vectors, colored by their respective motion type.}
    \label{fig:contexts}
\end{figure}

We then color this low-dimensional projection of the context vector space according to the type of motion performed.
Due to the large variety of different motion, we select to color motions of type walking, running, waving, wiping, mixing and squatting and use a combined color for all other motions.
The labels are obtained from the KIT Whole-Body Human Motion Database, which provides multiple labels for each motion record.
\added{If no labels are available or if motions have multiple conflicting labels, we discard this specific motion.}\footnote{\added{We would like to emphasize that the vast majority of motions that we used have exactly one label in each of these categories, which means that we only discard a small subset.}}
\autoref{fig:contexts} depicts this visualization for both models, motion-to-language and language-to-motion.

Both visualization exhibit a clear structure and contain clusters of motions of the same type.
Interestingly, the visualization of the motion-to-language model~(\autoref{fig:m2l_contexts}) has denser clusters with less variance and cleaner separation between clusters than the visualization of the language-to-motion model~(\autoref{fig:l2m_contexts}).
This makes intuitive sense since describing a motion in natural language has far more ambiguities compared to observing the motion directly.
Comparing running and walking motions, this observation is especially apparent:
In the motion-to-language case, the two types of motion are nicely separated wheres in the language-to-motion case the two types often fall into the same cluster (top-most green cluster in~\autoref{fig:l2m_contexts}).

Another interesting observation is that motions of type wiping and mixing (in the sense of mixing something in a bowl) are used interchangeably in both directions.
Since we do not include object information (to neither the model nor the human annotators that created the dataset), mixing and wiping appear to be the same.

\begin{figure}[h]
    \centering
    \includegraphics[width=0.7\textwidth,trim=1.5cm 0.5cm 1cm 1cm,clip=true]{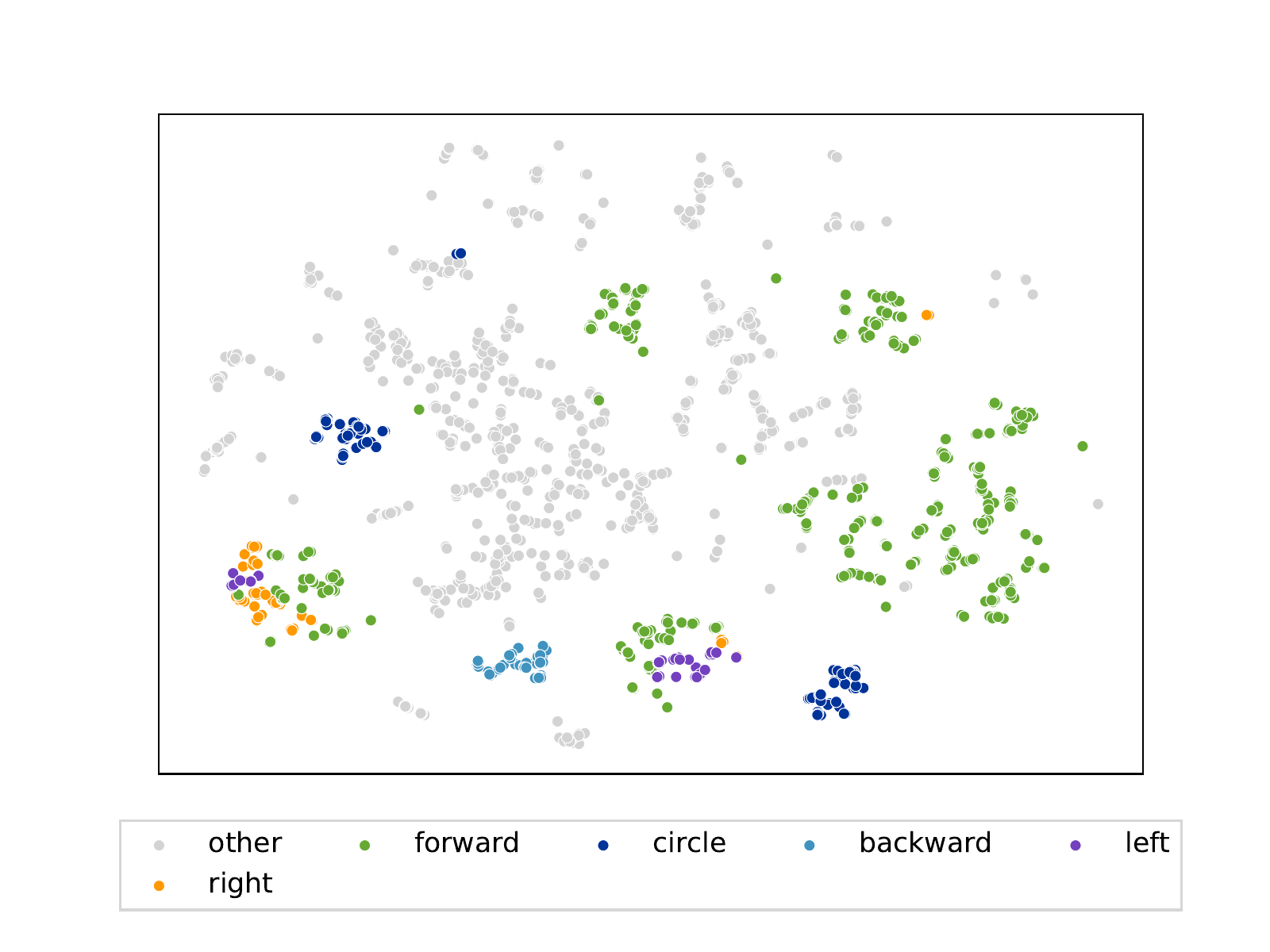}
    \caption{Visualization of contexts vectors associated with walking, colored by their respective direction.}
    \label{fig:m2l_contexts_walk}
\end{figure}

We also investigate the structure \emph{within} a given type of motion, in this case walking.
For this purpose, we use the same projection as in the previous motion-to-language visualization.
However, this time we color motions by their respective direction (left, right, forward, backward) and only those that are walking motions.
The resulting 2-dimensional t-SNE projection is depicted in \autoref{fig:m2l_contexts_walk}, which shows that walking motions that have the same direction are grouped in the same cluster.
However, the separation is often less clear than for motions of completely different type.

Overall, the analysis of the context vector space clearly suggests that our proposed models extract semantic meaning from the given input (for both cases, motion or language) and encode it into the context vector.
The decoder part of the models can then generate the correct sequence using this semantic representation.

\section{Conclusion}
\label{sec:conclusion}
In this paper, we proposed the use of deep recurrent neural networks in a sequence-to-sequence setting to learn a bidirectional mapping between human whole-body motion and descriptions in natural language.
We presented models that can be used to model each direction of the bidirectional mapping individually.
An important property of our proposed models is that they are probabilistic, allowing us to produce different candidate hypotheses and ranking them accordingly.
Additionally, our system makes minimal assumptions about both the natural language descriptions and human whole-body motions, requiring minimal preprocessing and no explicit motion segmentation into motion or action primitives or clusters thereof \emph{a-priori}.
Furthermore, each model makes use of a distributed representation, which is shared for all types of motions.

In our experiments, we clearly demonstrated the capabilities of our proposed system to generate rich and detailed descriptions of a large variety of different human whole-body motions.
Conversely, we showed that our model is capable of generating a similarly large variety of realistic human whole-body motion given a description thereof in natural language.
We quantified and reported the performance of our system using the \textsc{Bleu} score, which is well-known and frequently used in the context of machine translation.
We also presented results that indicate that each model successfully learns distributed and semantically meaningful latent representations of the given input to produce the desired output.

A limitation of our proposed system is that the input sequence needs to be encoded into a single vector (the context vector $\vec{c}$).
This becomes especially problematic as the sequence length increases.
To overcome this problem, \emph{attention mechanisms} have been proposed in the literature\citep{seq2seq-attn}.
Integrating such mechanisms in the future is likely going to improve the performance of our system.
Similarly, \emph{hierarchical RNNs} \citep{DBLP:journals/corr/JainZSS15, DBLP:conf/cvpr/DuWW15} have been proposed and were used successfully to model human motion.
Integrating these ideas into our system would likewise be an interesting experiment.

Increasing the size of the KIT Motion-Language Dataset~\citep{DBLP:journals/corr/PlappertMA16} is an important area of future work as well.
More data would allow us to use more complex models and reduces the risk of overfitting.
Additionally, including more complex motions in which a subject performs a sequence of distinct steps (e.g. when cooking) would allow for interesting experiments to further test the generalization capabilities of our proposed system by permutating the order of the steps.

Lastly, representing human whole-body motion using only the joint values of the kinematic model is insufficient.
In future work, we therefore intend to incorporate dynamic properties of the motion as well as contact information with the environment and objects that are involved in the execution of the motion. Such multi-contact motions have recently been studied by \cite{DBLP:conf/humanoids/ManderySJA15}.

\section*{Acknowledgments}
The research leading to these results has received funding from the H$^2$T at KIT, the German Research Foundation (Deutsche Forschungsgemeinschaft: DFG) under the Priority Program on Autonomous Learning (SPP 1527), and the European Union’s Horizon 2020 Research and Innovation Programme and Seventh Framework Programme under grant agreements No. 643666 (I-Support) and No. 611832 (WALK-MAN).

\bibliographystyle{SageH}
\bibliography{MotionLanguage}

\begin{thebibliography}{72}
\providecommand{\natexlab}[1]{#1}
\providecommand{\url}[1]{\texttt{#1}}
\providecommand{\urlprefix}{URL }
\expandafter\ifx\csname urlstyle\endcsname\relax
  \providecommand{\doi}[1]{DOI:\discretionary{}{}{}#1}\else
  \providecommand{\doi}{DOI:\discretionary{}{}{}\begingroup
  \urlstyle{rm}\Url}\fi

\bibitem[{Arie et~al.(2010)Arie, Endo, Jeong, Lee, Sugano and
  Tani}]{DBLP:conf/icann/ArieEJLST10}
Arie H, Endo T, Jeong S, Lee M, Sugano S and Tani J (2010) Integrative learning
  between language and action: {A} neuro-robotics experiment.
\newblock In: \emph{Artificial Neural Networks - {ICANN} 2010, 20th
  International Conference, Thessaloniki, Greece, September 15-18, 2010,
  Proceedings, Part {II}}. pp. 256--265.

\bibitem[{Azad et~al.(2007)Azad, Asfour and Dillmann}]{DBLP:conf/icra/AzadAD07}
Azad P, Asfour T and Dillmann R (2007) Toward an unified representation for
  imitation of human motion on humanoids.
\newblock In: \emph{2007 {IEEE} International Conference on Robotics and
  Automation, {ICRA} 2007, 10-14 April 2007, Roma, Italy}. pp. 2558--2563.

\bibitem[{Ba et~al.(2016)Ba, Kiros and Hinton}]{layernorm}
Ba LJ, Kiros R and Hinton GE (2016) Layer normalization.
\newblock \emph{CoRR} abs/1607.06450.

\bibitem[{Bahdanau et~al.(2014)Bahdanau, Cho and Bengio}]{seq2seq-attn}
Bahdanau D, Cho K and Bengio Y (2014) Neural machine translation by jointly
  learning to align and translate.
\newblock \emph{CoRR} abs/1409.0473.

\bibitem[{Bengio et~al.(2012)Bengio, Boulanger{-}Lewandowski and
  Pascanu}]{gradient_clipping}
Bengio Y, Boulanger{-}Lewandowski N and Pascanu R (2012) Advances in optimizing
  recurrent networks.
\newblock \emph{CoRR} abs/1212.0901.

\bibitem[{Billard et~al.(2008)Billard, Calinon, Dillmann and
  Schaal}]{DBLP:reference/robo/BillardCDS08}
Billard A, Calinon S, Dillmann R and Schaal S (2008) Robot programming by
  demonstration.
\newblock In: \emph{Springer Handbook of Robotics}. pp. 1371--1394.

\bibitem[{B{\"{u}}tepage et~al.(2017)B{\"{u}}tepage, Black, Kragic and
  Kjellstr{\"{o}}m}]{DBLP:journals/corr/ButepageBKK17}
B{\"{u}}tepage J, Black MJ, Kragic D and Kjellstr{\"{o}}m H (2017) Deep
  representation learning for human motion prediction and classification.
\newblock \emph{CoRR} abs/1702.07486.

\bibitem[{Calinon et~al.(2007)Calinon, Guenter and
  Billard}]{DBLP:journals/tsmc/CalinonGB07}
Calinon S, Guenter F and Billard A (2007) On learning, representing, and
  generalizing a task in a humanoid robot.
\newblock \emph{{IEEE} Trans. Systems, Man, and Cybernetics, Part {B}} 37(2):
  286--298.

\bibitem[{Cho et~al.(2014)Cho, van Merrienboer, Bahdanau and Bengio}]{gru1}
Cho K, van Merrienboer B, Bahdanau D and Bengio Y (2014) On the properties of
  neural machine translation: Encoder-decoder approaches.
\newblock \emph{CoRR} abs/1409.1259.

\bibitem[{Chung et~al.(2014)Chung, G{\"{u}}l{\c{c}}ehre, Cho and Bengio}]{gru2}
Chung J, G{\"{u}}l{\c{c}}ehre {\c{C}}, Cho K and Bengio Y (2014) Empirical
  evaluation of gated recurrent neural networks on sequence modeling.
\newblock \emph{CoRR} abs/1412.3555.

\bibitem[{Cooijmans et~al.(2016)Cooijmans, Ballas, Laurent and
  Courville}]{batchnorm_rnn}
Cooijmans T, Ballas N, Laurent C and Courville AC (2016) Recurrent batch
  normalization.
\newblock \emph{CoRR} abs/1603.09025.

\bibitem[{Dillmann et~al.(2000)Dillmann, Rogalla, Ehrenmann, Zollner and
  Bordegoni}]{dillmann2000learning}
Dillmann R, Rogalla O, Ehrenmann M, Zollner R and Bordegoni M (2000) Learning
  robot behaviour and skills based on human demonstration and advice: the
  machine learning paradigm.
\newblock In: \emph{ROBOTICS RESEARCH-INTERNATIONAL SYMPOSIUM-}, volume~9. pp.
  229--238.

\bibitem[{Donahue et~al.(2015)Donahue, Hendricks, Guadarrama, Rohrbach,
  Venugopalan, Darrell and Saenko}]{DBLP:conf/cvpr/DonahueHGRVDS15}
Donahue J, Hendricks LA, Guadarrama S, Rohrbach M, Venugopalan S, Darrell T and
  Saenko K (2015) Long-term recurrent convolutional networks for visual
  recognition and description.
\newblock In: \emph{{IEEE} Conference on Computer Vision and Pattern
  Recognition, {CVPR} 2015, Boston, MA, USA, June 7-12, 2015}. pp. 2625--2634.

\bibitem[{Dozat(2015)}]{nadam}
Dozat T (2015) Incorporating nesterov momentum into adam.
\newblock Technical report, Stanford University, Tech. Rep., 2015.

\bibitem[{Du et~al.(2015)Du, Wang and Wang}]{DBLP:conf/cvpr/DuWW15}
Du Y, Wang W and Wang L (2015) Hierarchical recurrent neural network for
  skeleton based action recognition.
\newblock In: \emph{{IEEE} Conference on Computer Vision and Pattern
  Recognition, {CVPR} 2015, Boston, MA, USA, June 7-12, 2015}. pp. 1110--1118.

\bibitem[{Field et~al.(2011)Field, Pan, Stirling and Naghdy}]{field2011mocap}
Field M, Pan Z, Stirling D and Naghdy F (2011) Human motion capture sensors and
  analysis in robotics 38(2): 163--171.

\bibitem[{Fragkiadaki et~al.(2015)Fragkiadaki, Levine and
  Malik}]{DBLP:journals/corr/FragkiadakiLM15}
Fragkiadaki K, Levine S and Malik J (2015) Recurrent network models for
  kinematic tracking.
\newblock \emph{CoRR} abs/1508.00271.

\bibitem[{Gal(2015)}]{dropout_rnn}
Gal Y (2015) A theoretically grounded application of dropout in recurrent
  neural networks.
\newblock \emph{arXiv preprint arXiv:1512.05287} .

\bibitem[{Gers et~al.(2000)Gers, Schmidhuber and Cummins}]{lstm2}
Gers FA, Schmidhuber J and Cummins F (2000) Learning to forget: Continual
  prediction with lstm.
\newblock \emph{Neural computation} 12(10): 2451--2471.

\bibitem[{Glorot et~al.(2011)Glorot, Bordes and
  Bengio}]{DBLP:conf/icml/GlorotBB11}
Glorot X, Bordes A and Bengio Y (2011) Domain adaptation for large-scale
  sentiment classification: {A} deep learning approach.
\newblock In: \emph{Proceedings of the 28th International Conference on Machine
  Learning, {ICML} 2011, Bellevue, Washington, USA, June 28 - July 2, 2011}.
  pp. 513--520.

\bibitem[{Goodfellow et~al.(2016)Goodfellow, Bengio and
  Courville}]{Goodfellow-et-al-2016-Book}
Goodfellow I, Bengio Y and Courville A (2016) Deep learning.
\newblock Book in preparation for MIT Press.

\bibitem[{Graves(2013)}]{graves_rnns}
Graves A (2013) Generating sequences with recurrent neural networks.
\newblock \emph{CoRR} abs/1308.0850.

\bibitem[{Graves et~al.(2013)Graves, Mohamed and
  Hinton}]{DBLP:conf/icassp/GravesMH13}
Graves A, Mohamed A and Hinton GE (2013) Speech recognition with deep recurrent
  neural networks.
\newblock In: \emph{{IEEE} International Conference on Acoustics, Speech and
  Signal Processing, {ICASSP} 2013, Vancouver, BC, Canada, May 26-31, 2013}.
  pp. 6645--6649.

\bibitem[{Graves and Schmidhuber(2005)}]{brnn}
Graves A and Schmidhuber J (2005) Framewise phoneme classification with
  bidirectional {LSTM} and other neural network architectures.
\newblock \emph{Neural Networks} 18(5-6): 602--610.

\bibitem[{Gu et~al.(2016)Gu, Holly, Lillicrap and
  Levine}]{DBLP:journals/corr/GuHLL16}
Gu S, Holly E, Lillicrap TP and Levine S (2016) Deep reinforcement learning for
  robotic manipulation.
\newblock \emph{CoRR} abs/1610.00633.

\bibitem[{Hassani and Lee(2016)}]{natural_language_descr_survey}
Hassani K and Lee W (2016) Visualizing natural language descriptions: {A}
  survey.
\newblock \emph{{ACM} Comput. Surv.} 49(1): 17.

\bibitem[{He et~al.(2015)He, Zhang, Ren and Sun}]{DBLP:journals/corr/HeZRS15}
He K, Zhang X, Ren S and Sun J (2015) Deep residual learning for image
  recognition.
\newblock \emph{CoRR} abs/1512.03385.

\bibitem[{Herzog et~al.(2008)Herzog, Ude and
  Kr{\"{u}}ger}]{DBLP:conf/humanoids/HerzogUK08}
Herzog D, Ude A and Kr{\"{u}}ger V (2008) Motion imitation and recognition
  using parametric hidden markov models.
\newblock In: \emph{8th {IEEE-RAS} International Conference on Humanoid Robots,
  Humanoids 2008, Daejeon, South Korea, December 1-3, 2008}. pp. 339--346.

\bibitem[{Hochreiter and Schmidhuber(1997)}]{lstm1}
Hochreiter S and Schmidhuber J (1997) Long short-term memory.
\newblock \emph{Neural computation} 9(8): 1735--1780.

\bibitem[{Hodosh et~al.(2013)Hodosh, Young and
  Hockenmaier}]{DBLP:journals/jair/HodoshYH13}
Hodosh M, Young P and Hockenmaier J (2013) Framing image description as a
  ranking task: Data, models and evaluation metrics.
\newblock \emph{J. Artif. Intell. Res.} 47: 853--899.

\bibitem[{Huang et~al.(2001)Huang, Acero and Hon}]{huang2001spoken}
Huang X, Acero A and Hon HW (2001) \emph{Spoken language processing: A guide to
  theory, algorithm, and system development}.
\newblock 1st ed. edition. Englewood Cliffs, NJ, USA: Prentice Hall.

\bibitem[{Ioffe and Szegedy(2015)}]{batchnorm}
Ioffe S and Szegedy C (2015) Batch normalization: Accelerating deep network
  training by reducing internal covariate shift.
\newblock In: \emph{Proceedings of the 32nd International Conference on Machine
  Learning, {ICML} 2015, Lille, France, 6-11 July 2015}. pp. 448--456.

\bibitem[{Jain et~al.(2015)Jain, Zamir, Savarese and
  Saxena}]{DBLP:journals/corr/JainZSS15}
Jain A, Zamir AR, Savarese S and Saxena A (2015) Structural-rnn: Deep learning
  on spatio-temporal graphs.
\newblock \emph{CoRR} abs/1511.05298.

\bibitem[{Karpathy and Li(2015)}]{DBLP:conf/cvpr/KarpathyL15}
Karpathy A and Li F (2015) Deep visual-semantic alignments for generating image
  descriptions.
\newblock In: \emph{{IEEE} Conference on Computer Vision and Pattern
  Recognition, {CVPR} 2015, Boston, MA, USA, June 7-12, 2015}. pp. 3128--3137.

\bibitem[{Kingma and Ba(2014)}]{adam}
Kingma DP and Ba J (2014) Adam: {A} method for stochastic optimization.
\newblock \emph{CoRR} abs/1412.6980.

\bibitem[{Krizhevsky et~al.(2012)Krizhevsky, Sutskever and
  Hinton}]{DBLP:conf/nips/KrizhevskySH12}
Krizhevsky A, Sutskever I and Hinton GE (2012) Imagenet classification with
  deep convolutional neural networks.
\newblock In: \emph{Advances in Neural Information Processing Systems 25: 26th
  Annual Conference on Neural Information Processing Systems 2012. Proceedings
  of a meeting held December 3-6, 2012, Lake Tahoe, Nevada, United States.} pp.
  1106--1114.

\bibitem[{Kulic et~al.(2008)Kulic, Takano and
  Nakamura}]{DBLP:journals/ijrr/KulicTN08}
Kulic D, Takano W and Nakamura Y (2008) Incremental learning, clustering and
  hierarchy formation of whole body motion patterns using adaptive hidden
  markov chains.
\newblock \emph{I. J. Robotics Res.} 27(7): 761--784.

\bibitem[{Kuniyoshi et~al.(1994)Kuniyoshi, Inaba and Inoue}]{kuniyoshi1994}
Kuniyoshi Y, Inaba M and Inoue H (1994) Learning by watching: Extracting
  reusable task knowledge from visual observation of human performance.
\newblock \emph{IEEE Transactions on Robotics and Automation} 10: 799--822.

\bibitem[{LeCun et~al.(2015)LeCun, Bengio and Hinton}]{lecun2015deep}
LeCun Y, Bengio Y and Hinton G (2015) Deep learning.
\newblock \emph{Nature} 521(7553): 436--444.

\bibitem[{Levine et~al.(2015)Levine, Finn, Darrell and
  Abbeel}]{DBLP:journals/corr/LevineFDA15}
Levine S, Finn C, Darrell T and Abbeel P (2015) End-to-end training of deep
  visuomotor policies.
\newblock \emph{CoRR} abs/1504.00702.

\bibitem[{Levine et~al.(2016)Levine, Pastor, Krizhevsky and
  Quillen}]{DBLP:journals/corr/LevinePKQ16}
Levine S, Pastor P, Krizhevsky A and Quillen D (2016) Learning hand-eye
  coordination for robotic grasping with deep learning and large-scale data
  collection.
\newblock \emph{CoRR} abs/1603.02199.

\bibitem[{Maaten and Hinton(2008)}]{tsne}
Maaten Lvd and Hinton G (2008) Visualizing high-dimensional data using {t-SNE}.
\newblock \emph{Journal of Machine Learning Research} 9(Nov): 2579--2605.

\bibitem[{Mandery et~al.(2015{\natexlab{a}})Mandery, Sol, J{\"{o}}chner and
  Asfour}]{DBLP:conf/humanoids/ManderySJA15}
Mandery C, Sol JB, J{\"{o}}chner M and Asfour T (2015{\natexlab{a}}) Analyzing
  whole-body pose transitions in multi-contact motions.
\newblock In: \emph{15th {IEEE-RAS} International Conference on Humanoid
  Robots, Humanoids 2015, Seoul, South Korea, November 3-5, 2015}. pp.
  1020--1027.

\bibitem[{Mandery et~al.(2015{\natexlab{b}})Mandery, Terlemez, Do, Vahrenkamp
  and Asfour}]{DBLP:conf/icar/ManderyTDVA15}
Mandery C, Terlemez {\"{O}}, Do M, Vahrenkamp N and Asfour T
  (2015{\natexlab{b}}) The {KIT} whole-body human motion database.
\newblock In: \emph{International Conference on Advanced Robotics, {ICAR} 2015,
  Istanbul, Turkey, July 27-31, 2015}. pp. 329--336.

\bibitem[{Mandery et~al.(2016)Mandery, Terlemez, Do, Vahrenkamp and
  Asfour}]{DBLP:journals/trob/ManderyTDVA16}
Mandery C, Terlemez {\"{O}}, Do M, Vahrenkamp N and Asfour T (2016) Unifying
  representations and large-scale whole-body motion databases for studying
  human motion.
\newblock \emph{{IEEE} Trans. Robotics} 32(4): 796--809.

\bibitem[{Medina et~al.(2012)Medina, Shelley, Lee, Takano and
  Hirche}]{DBLP:conf/ro-man/MedinaSLTH12}
Medina JR, Shelley M, Lee D, Takano W and Hirche S (2012) Towards interactive
  physical robotic assistance: Parameterizing motion primitives through natural
  language.
\newblock In: \emph{The 21st {IEEE} International Symposium on Robot and Human
  Interactive Communication, {IEEE} {RO-MAN} 2012, Paris, France, September
  9-13, 2012}. pp. 1097--1102.

\bibitem[{Mikolov et~al.(2013)Mikolov, Sutskever, Chen, Corrado and
  Dean}]{word_embeddings1}
Mikolov T, Sutskever I, Chen K, Corrado GS and Dean J (2013) Distributed
  representations of words and phrases and their compositionality.
\newblock In: \emph{Advances in Neural Information Processing Systems 26: 27th
  Annual Conference on Neural Information Processing Systems 2013. Proceedings
  of a meeting held December 5-8, 2013, Lake Tahoe, Nevada, United States.} pp.
  3111--3119.

\bibitem[{Mordatch et~al.(2015)Mordatch, Lowrey, Andrew, Popovic and
  Todorov}]{DBLP:conf/nips/MordatchLAPT15}
Mordatch I, Lowrey K, Andrew G, Popovic Z and Todorov E (2015) Interactive
  control of diverse complex characters with neural networks.
\newblock In: \emph{Advances in Neural Information Processing Systems 28:
  Annual Conference on Neural Information Processing Systems 2015, December
  7-12, 2015, Montreal, Quebec, Canada}. pp. 3132--3140.

\bibitem[{Ogata et~al.(2007{\natexlab{a}})Ogata, Matsumoto, Tani, Komatani and
  Okuno}]{DBLP:conf/icra/OgataMTKO07}
Ogata T, Matsumoto S, Tani J, Komatani K and Okuno HG (2007{\natexlab{a}})
  Human-robot cooperation using quasi-symbols generated by {RNNPB} model.
\newblock In: \emph{2007 {IEEE} International Conference on Robotics and
  Automation, {ICRA} 2007, 10-14 April 2007, Roma, Italy}. pp. 2156--2161.

\bibitem[{Ogata et~al.(2007{\natexlab{b}})Ogata, Murase, Tani, Komatani and
  Okuno}]{DBLP:conf/iros/OgataMTKO07}
Ogata T, Murase M, Tani J, Komatani K and Okuno HG (2007{\natexlab{b}}) Two-way
  translation of compound sentences and arm motions by recurrent neural
  networks.
\newblock In: \emph{2007 {IEEE/RSJ} International Conference on Intelligent
  Robots and Systems, October 29 - November 2, 2007, Sheraton Hotel and Marina,
  San Diego, California, {USA}}. pp. 1858--1863.

\bibitem[{Ogata and Okuno(2013)}]{DBLP:conf/riiss/OgataO13}
Ogata T and Okuno HG (2013) Integration of behaviors and languages with a
  hierarchal structure self-organized in a neuro-dynamical model.
\newblock In: \emph{2013 {IEEE} Workshop on Robotic Intelligence In
  Informationally Structured Space, RiiSS 2013, Singapore, April 16-19, 2013}.
  pp. 89--95.

\bibitem[{Papineni et~al.(2002)Papineni, Roukos, Ward and
  Zhu}]{DBLP:conf/acl/PapineniRWZ02}
Papineni K, Roukos S, Ward T and Zhu W (2002) Bleu: a method for automatic
  evaluation of machine translation.
\newblock In: \emph{Proceedings of the 40th Annual Meeting of the Association
  for Computational Linguistics, July 6-12, 2002, Philadelphia, PA, {USA.}} pp.
  311--318.

\bibitem[{Plappert et~al.(2016)Plappert, Mandery and
  Asfour}]{DBLP:journals/corr/PlappertMA16}
Plappert M, Mandery C and Asfour T (2016) The {KIT} motion-language dataset.
\newblock \emph{CoRR} abs/1607.03827.

\bibitem[{Schaal(2006)}]{Schaal2006}
Schaal S (2006) \emph{Dynamic Movement Primitives -A Framework for Motor
  Control in Humans and Humanoid Robotics}.
\newblock Tokyo: Springer Tokyo, pp. 261--280.

\bibitem[{Srivastava et~al.(2014)Srivastava, Hinton, Krizhevsky, Sutskever and
  Salakhutdinov}]{dropout}
Srivastava N, Hinton GE, Krizhevsky A, Sutskever I and Salakhutdinov R (2014)
  Dropout: a simple way to prevent neural networks from overfitting.
\newblock \emph{Journal of Machine Learning Research} 15(1): 1929--1958.

\bibitem[{Sugita and Tani(2005)}]{DBLP:journals/adb/SugitaT05}
Sugita Y and Tani J (2005) Learning semantic combinatoriality from the
  interaction between linguistic and behavioral processes.
\newblock \emph{Adaptive Behaviour} 13(1): 33--52.

\bibitem[{Sutskever et~al.(2014)Sutskever, Vinyals and Le}]{seq2seq1}
Sutskever I, Vinyals O and Le QV (2014) Sequence to sequence learning with
  neural networks.
\newblock In: \emph{Advances in Neural Information Processing Systems 27:
  Annual Conference on Neural Information Processing Systems 2014, December
  8-13 2014, Montreal, Quebec, Canada}. pp. 3104--3112.

\bibitem[{Takano et~al.(2007)Takano, Kulic and
  Nakamura}]{DBLP:conf/iros/TakanoKN07}
Takano W, Kulic D and Nakamura Y (2007) Interactive topology formation of
  linguistic space and motion space.
\newblock In: \emph{2007 {IEEE/RSJ} International Conference on Intelligent
  Robots and Systems, October 29 - November 2, 2007, Sheraton Hotel and Marina,
  San Diego, California, {USA}}. pp. 1416--1422.

\bibitem[{Takano et~al.(2016)Takano, Kusajima and
  Nakamura}]{DBLP:journals/nn/TakanoKN16}
Takano W, Kusajima I and Nakamura Y (2016) Generating action descriptions from
  statistically integrated representations of human motions and sentences.
\newblock \emph{Neural Networks} 80: 1--8.

\bibitem[{Takano and Nakamura(2008)}]{DBLP:conf/humanoids/TakanoN08}
Takano W and Nakamura Y (2008) Integrating whole body motion primitives and
  natural language for humanoid robots.
\newblock In: \emph{8th {IEEE-RAS} International Conference on Humanoid Robots,
  Humanoids 2008, Daejeon, South Korea, December 1-3, 2008}. pp. 708--713.

\bibitem[{Takano and Nakamura(2009)}]{DBLP:conf/icra/TakanoN09}
Takano W and Nakamura Y (2009) Statistically integrated semiotics that enables
  mutual inference between linguistic and behavioral symbols for humanoid
  robots.
\newblock In: \emph{2009 {IEEE} International Conference on Robotics and
  Automation, {ICRA} 2009, Kobe, Japan, May 12-17, 2009}. pp. 646--652.

\bibitem[{Takano and Nakamura(2012)}]{DBLP:conf/icra/TakanoN12}
Takano W and Nakamura Y (2012) Bigram-based natural language model and
  statistical motion symbol model for scalable language of humanoid robots.
\newblock In: \emph{{IEEE} International Conference on Robotics and Automation,
  {ICRA} 2012, 14-18 May, 2012, St. Paul, Minnesota, {USA}}. pp. 1232--1237.

\bibitem[{Takano and
  Nakamura(2015{\natexlab{a}})}]{DBLP:journals/ijrr/TakanoN15}
Takano W and Nakamura Y (2015{\natexlab{a}}) Statistical mutual conversion
  between whole body motion primitives and linguistic sentences for human
  motions.
\newblock \emph{I. J. Robotics Res.} 34(10): 1314--1328.

\bibitem[{Takano and
  Nakamura(2015{\natexlab{b}})}]{DBLP:journals/ras/TakanoN15}
Takano W and Nakamura Y (2015{\natexlab{b}}) Symbolically structured database
  for human whole body motions based on association between motion symbols and
  motion words.
\newblock \emph{Robotics and Autonomous Systems} 66: 75--85.

\bibitem[{Takano et~al.(2006)Takano, Yamane, Sugihara, Yamamoto and
  Nakamura}]{DBLP:conf/icra/TakanoYSYN06}
Takano W, Yamane K, Sugihara T, Yamamoto K and Nakamura Y (2006) Primitive
  communication based on motion recognition and generation with hierarchical
  mimesis model.
\newblock In: \emph{Proceedings of the 2006 {IEEE} International Conference on
  Robotics and Automation, {ICRA} 2006, May 15-19, 2006, Orlando, Florida,
  {USA}}. pp. 3602--3609.

\bibitem[{Taylor and Hinton(2009)}]{DBLP:conf/icml/TaylorH09}
Taylor GW and Hinton GE (2009) Factored conditional restricted boltzmann
  machines for modeling motion style.
\newblock In: \emph{Proceedings of the 26th Annual International Conference on
  Machine Learning, {ICML} 2009, Montreal, Quebec, Canada, June 14-18, 2009}.
  pp. 1025--1032.

\bibitem[{Taylor et~al.(2006)Taylor, Hinton and
  Roweis}]{DBLP:conf/nips/TaylorHR06}
Taylor GW, Hinton GE and Roweis ST (2006) Modeling human motion using binary
  latent variables.
\newblock In: \emph{Advances in Neural Information Processing Systems 19,
  Proceedings of the Twentieth Annual Conference on Neural Information
  Processing Systems, Vancouver, British Columbia, Canada, December 4-7, 2006}.
  pp. 1345--1352.

\bibitem[{Terlemez et~al.(2014)Terlemez, Ulbrich, Mandery, Do, Vahrenkamp and
  Asfour}]{DBLP:conf/humanoids/TerlemezUMDVA14}
Terlemez {\"{O}}, Ulbrich S, Mandery C, Do M, Vahrenkamp N and Asfour T (2014)
  Master motor map {(MMM)} - framework and toolkit for capturing, representing,
  and reproducing human motion on humanoid robots.
\newblock In: \emph{14th {IEEE-RAS} International Conference on Humanoid
  Robots, Humanoids 2014, Madrid, Spain, November 18-20, 2014}. pp. 894--901.

\bibitem[{Venugopalan et~al.(2015)Venugopalan, Xu, Donahue, Rohrbach, Mooney
  and Saenko}]{DBLP:conf/naacl/VenugopalanXDRM15}
Venugopalan S, Xu H, Donahue J, Rohrbach M, Mooney RJ and Saenko K (2015)
  Translating videos to natural language using deep recurrent neural networks.
\newblock In: \emph{{NAACL} {HLT} 2015, The 2015 Conference of the North
  American Chapter of the Association for Computational Linguistics: Human
  Language Technologies, Denver, Colorado, USA, May 31 - June 5, 2015}. pp.
  1494--1504.

\bibitem[{Vinyals et~al.(2015)Vinyals, Toshev, Bengio and
  Erhan}]{DBLP:conf/cvpr/VinyalsTBE15}
Vinyals O, Toshev A, Bengio S and Erhan D (2015) Show and tell: {A} neural
  image caption generator.
\newblock In: \emph{{IEEE} Conference on Computer Vision and Pattern
  Recognition, {CVPR} 2015, Boston, MA, USA, June 7-12, 2015}. pp. 3156--3164.

\bibitem[{Werbos(1990)}]{bptt}
Werbos PJ (1990) Backpropagation through time: what it does and how to do it.
\newblock \emph{Proceedings of the IEEE} 78(10): 1550--1560.

\bibitem[{Wu et~al.(2016)Wu, Schuster, Chen, Le, Norouzi, Macherey, Krikun,
  Cao, Gao, Macherey, Klingner, Shah, Johnson, Liu, Kaiser, Gouws, Kato, Kudo,
  Kazawa, Stevens, Kurian, Patil, Wang, Young, Smith, Riesa, Rudnick, Vinyals,
  Corrado, Hughes and Dean}]{DBLP:journals/corr/WuSCLNMKCGMKSJL16}
Wu Y, Schuster M, Chen Z, Le QV, Norouzi M, Macherey W, Krikun M, Cao Y, Gao Q,
  Macherey K, Klingner J, Shah A, Johnson M, Liu X, Kaiser L, Gouws S, Kato Y,
  Kudo T, Kazawa H, Stevens K, Kurian G, Patil N, Wang W, Young C, Smith J,
  Riesa J, Rudnick A, Vinyals O, Corrado G, Hughes M and Dean J (2016) Google's
  neural machine translation system: Bridging the gap between human and machine
  translation.
\newblock \emph{CoRR} abs/1609.08144.

\end{thebibliography}

\clearpage
\appendix
\onecolumn
\section{Detailed motion-to-language model architecture}
\label{appendix:m2l_model}
\begin{figure}[h]
    \centering
    \includegraphics[width=0.7\textwidth]{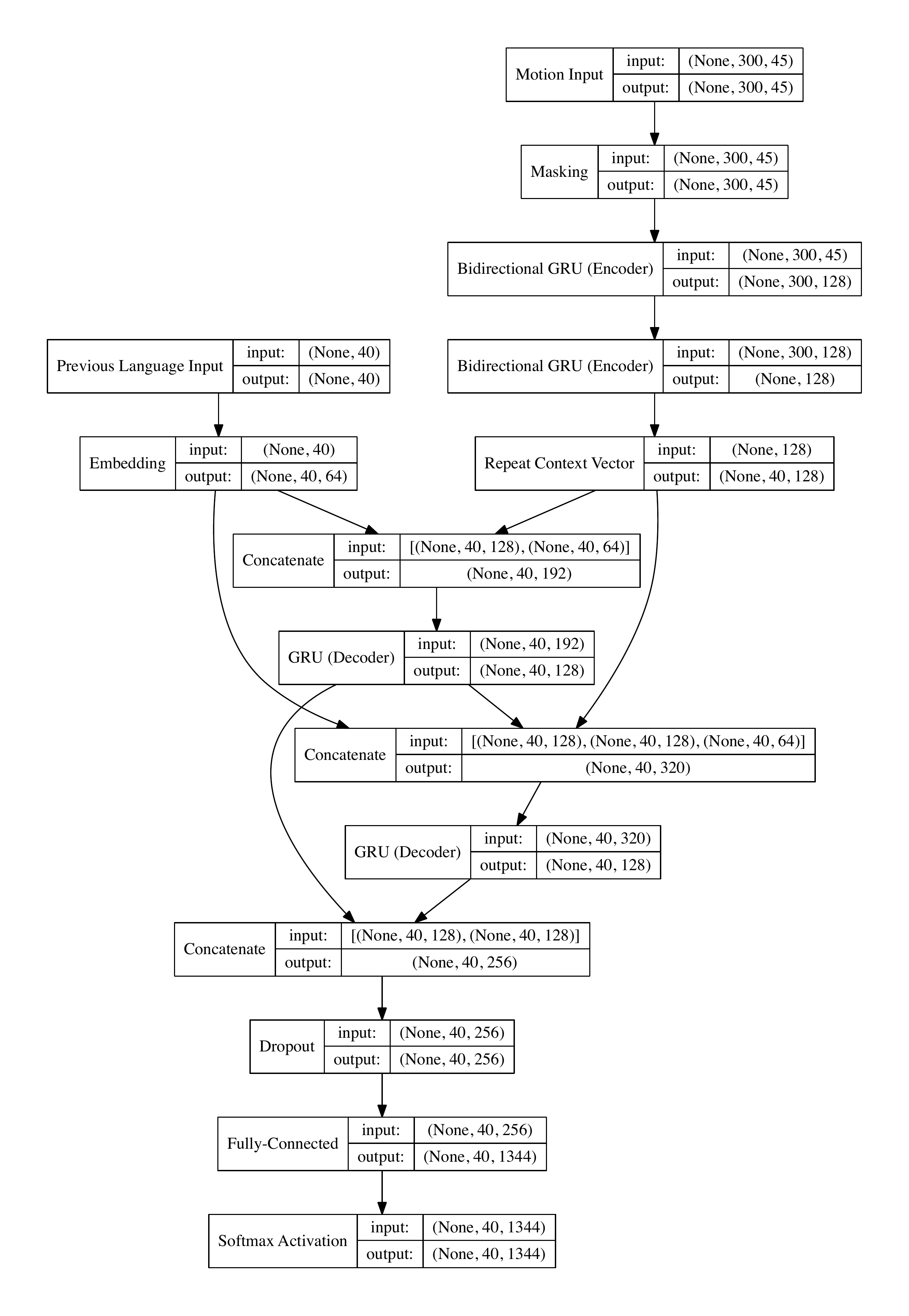}
    \caption{The detailed architecture of the motion-to-language model. Each node specifies the shape of its input tensor and output tensor. The first dimension is the batch size (128 in this case) followed by the time dimension for 3-dimensional tensors (padded to 300 timesteps at 10 Hz for the motion case and padded to 41 words for the language case). The last dimension is the feature dimension, of which the meaning depends on the specific modality and layer.}
\end{figure}
\clearpage

\section{Detailed language-to-motion model architecture}
\label{appendix:l2m_model}
\begin{figure}[h]
    \centering
    \includegraphics[width=\textwidth]{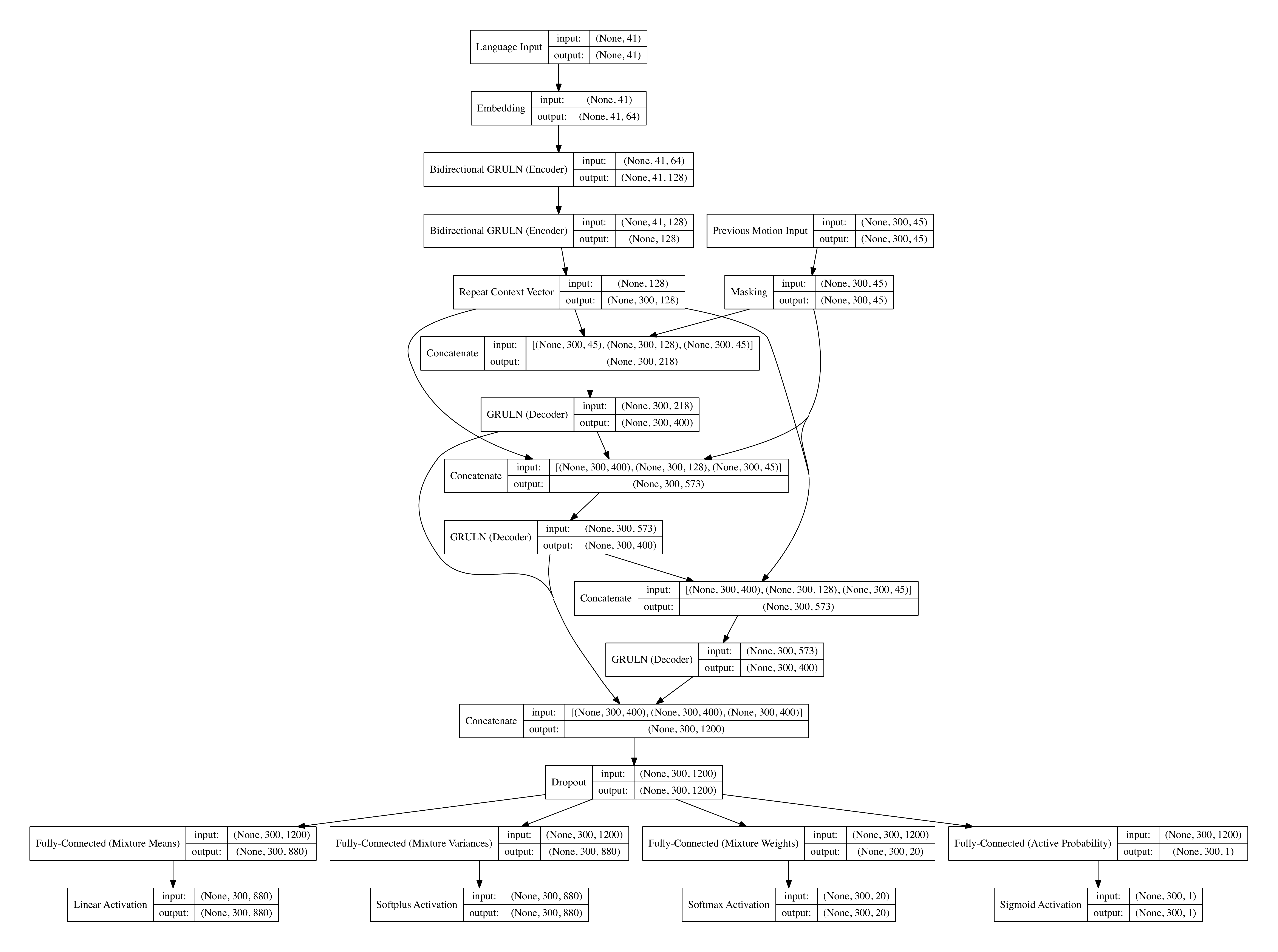}
    \caption{The detailed architecture of the language-to-motion model. Each node specifies the shape of its input tensor and output tensor. The first dimension is the batch size (128 in this case) followed by the time dimension for 3-dimensional tensors (padded to 300 timesteps at 10 Hz for the motion case and padded to 41 words for the language case). The last dimension is the feature dimension, of which the meaning depends on the specific modality and layer.}
\end{figure}

\end{document}